\documentclass[letterpaper,journal]{IEEEtran}

\usepackage[nocompress]{cite}
\usepackage{amsmath}
\usepackage{algorithm}
\usepackage{algpseudocode}
\usepackage{amssymb}
\usepackage{bm}
\usepackage{graphicx}
\usepackage{booktabs}
\usepackage{multirow}
\usepackage{makecell}
\usepackage{array}
\usepackage{tabularx}
\usepackage{adjustbox}
\usepackage{enumitem}
\usepackage{colortbl}
\usepackage{xcolor}
\usepackage{tcolorbox}
\usepackage{stfloats}
\usepackage{placeins}
\usepackage{pifont}          
\usepackage[colorlinks]{hyperref}

\definecolor{reference}{RGB}{55,126,168}
\newtcolorbox{promptbox}[1]{
  colback=reference!3,
  colframe=reference!38,
  colbacktitle=reference!12,
  coltitle=black,
  boxrule=0.45pt,
  arc=1.5pt,
  left=5pt,
  right=5pt,
  top=4pt,
  bottom=4pt,
  before skip=5pt,
  after skip=5pt,
  fonttitle=\bfseries\small,
  fontupper=\small,
  title={#1}
}
\AtBeginDocument{
  \hypersetup{
    linkcolor=reference,
    anchorcolor=reference,
    citecolor=reference,
    filecolor=reference,
    urlcolor=reference,
    pdftitle={Traffic-VL: Reinforced Vision-Language Models for Generalizable End-to-End Traffic Signal Control},
    pdfauthor={Xingchen Zou, Maonan Wang, Kemou Jiang, Jie Li, Yiqi Chen, Xi Chen, Hai Yang, and Yuxuan Liang}
  }
}

\newcolumntype{L}[1]{>{\raggedright\arraybackslash}p{#1}}
\newcolumntype{C}[1]{>{\centering\arraybackslash}p{#1}}
\newcolumntype{R}[1]{>{\raggedleft\arraybackslash}p{#1}}

\begin{document}
\bstctlcite{IEEEreferenceFormatting}

\title{Traffic-VL: Reinforced Vision-Language Models for Generalizable End-to-End Traffic Signal Control}

\author{Xingchen Zou, Maonan Wang, Kemou Jiang, Jie Li, Yiqi Chen, Xi Chen, Hai Yang, and Yuxuan Liang%
\thanks{This work is supported by the National Natural Science Foundation of China (No. 62402414), the Guangdong Basic and Applied Basic Research Foundation (No. 2025A1515011994), Guangzhou Municipal Science and Technology Project (No. 2023A03J0011), the Guangzhou Industrial Information and Intelligent Key Laboratory Project (No. 2024A03J0628), and Guangdong Provincial Key Lab of Integrated Communication, Sensing and Computation for Ubiquitous Internet of Things (No. 2023B1212010007).}%
\thanks{Xingchen Zou and Yuxuan Liang are with The Hong Kong University of Science and Technology (Guangzhou), Guangzhou, China. Corresponding author: Yuxuan Liang (e-mail: yuxliang@outlook.com).}%
\thanks{Maonan Wang and Xi Chen are with the Department of Mechanical and Automation Engineering, The Chinese University of Hong Kong, Hong Kong, China.}%
\thanks{Kemou Jiang is with Beihang University, Beijing, China.}%
\thanks{Jie Li and Yiqi Chen are with PCITECH, Guangzhou, China.}%
\thanks{Hai Yang is with the Department of Civil and Environmental Engineering, The Hong Kong University of Science and Technology, Clear Water Bay, Kowloon, Hong Kong, China.}%
\thanks{This article extends our conference paper ``Traffic-R1: Reinforced LLMs Bring Human-Like
Reasoning to Traffic Signal Control Systems'' (the 64th Annual Meeting of the
Association for Computational Linguistics)~\cite{zou2025trafficr1}.}}

\maketitle

\begin{abstract}
Existing traffic signal control methods typically rely on structured traffic states produced by dedicated perception pipelines. Emerging visual approaches often use idealized simulator images and convert them into intermediate representations before control, limiting their ability to directly exploit roadside visual observations. To bridge both the visual-domain and perception-to-decision gaps, we introduce SynTra-V and Traffic-VL. SynTra-V converts physics-based traffic simulations into photorealistic multi-view surveillance imagery across diverse road topologies and sensing conditions, and automatic semantic auditing checks that control-critical geometry, vehicle movements, and lane semantics match the simulator state. Traffic-VL is an eight-billion-parameter vision-language model that directly maps synchronized multi-view images and natural-language phase descriptions to signal decisions without intermediate structured states. Building on Traffic-R1's two-stage reinforced reasoning framework, Traffic-VL adapts its vision encoder via offline reinforcement learning on SynTra-V and freezes it during online visual stepwise trajectory policy optimization in closed-loop simulation. Experiments show that Traffic-VL matches structured-state Traffic-R1 while outperforming image-based controllers and general-purpose vision-language models, including Gemini 3 Pro. Evaluations on unseen topologies, degraded sensing conditions, real surveillance footage, and a three-week deployment across ten intersections further demonstrate robust visual generalization and effective constrained online control. 
\end{abstract}


\section{Introduction}

\begin{figure}[t]
  \centering
  \includegraphics[width=0.95\linewidth]{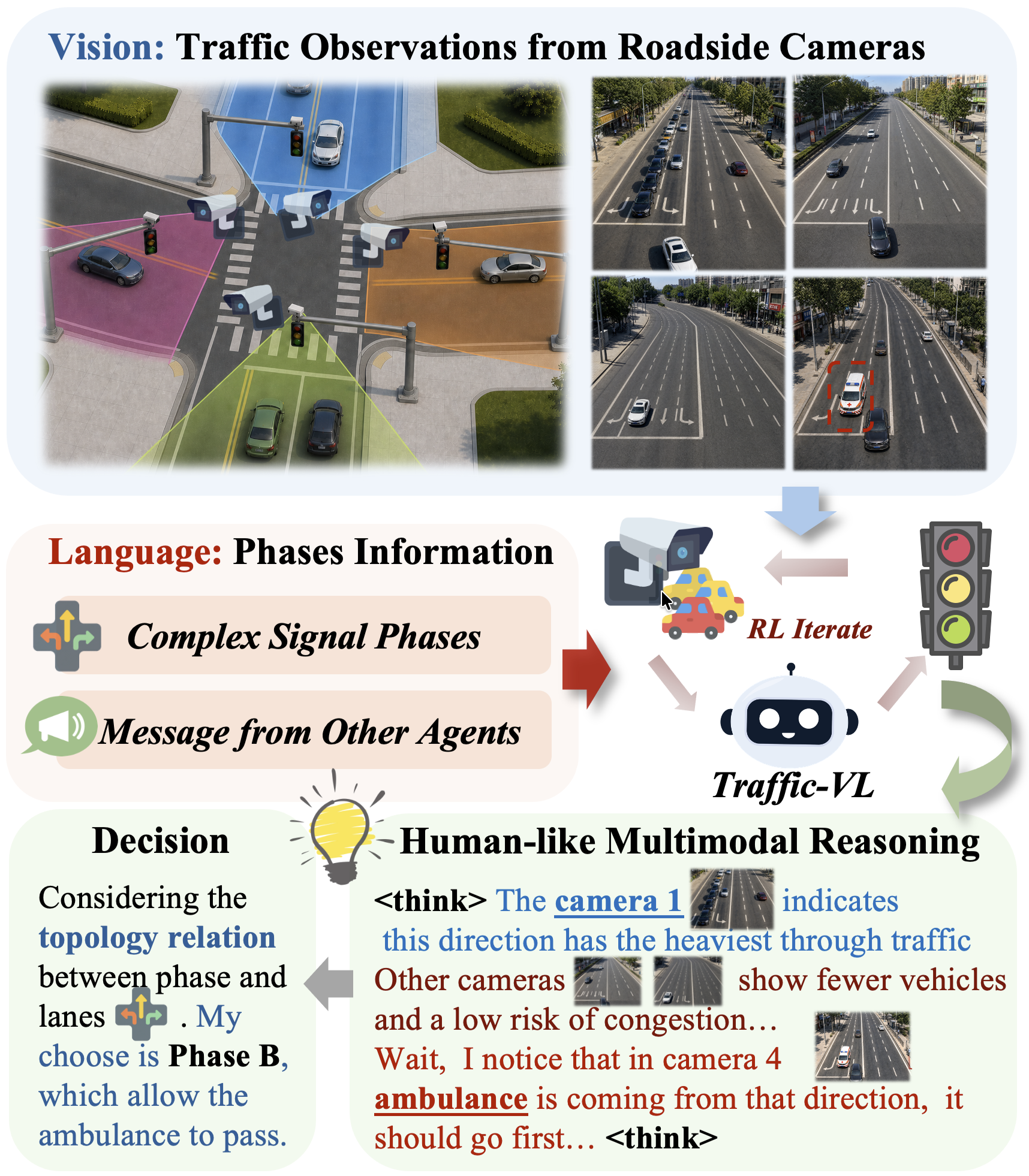}
  \caption{Traffic-VL maps multi-view roadside images and natural-language phase descriptions directly to traffic signal decisions.}
  \label{fig:teaser}
\end{figure}

\IEEEPARstart{A}{s} a critical urban public service, traffic signal control (TSC) directly affects
transportation efficiency, energy consumption, and emissions~\cite{xiao2026intelligent,hassan2026comprehensive}.
Effective control must respond to changing traffic demand across intersections with
different road layouts, sensing conditions, and signal plans. This diversity makes
generalization beyond individually engineered intersections a central challenge.

Reinforcement learning (RL) has advanced data-driven TSC by learning phase-selection
policies from lane-level traffic measurements~\cite{wei2019presslight,wei2019colight}.
More recently, large language models (LLMs) have enabled reasoning over textual traffic
observations, as in LLMLight~\cite{lai2025llmlight} and CoLLMLight~\cite{yuan2026collmlight}.
Our conference work Traffic-R1~\cite{zou2025trafficr1} extends this foundation-model
paradigm with reinforced reasoning for zero-shot and out-of-distribution control.
However, in these pipelines, the upstream state interface still determines which evidence
reaches the policy. For example, two queues of equal length may require different
priorities if one contains an ambulance; count-only states cannot express this distinction.
This limitation motivates keeping camera evidence available throughout phase reasoning,
allowing the policy to use details that a fixed upstream representation may omit.

Visual TSC methods such as VLMLight~\cite{wang2025vlmlight} and
ReasonLight~\cite{pang2026reasonlight} use vision-language models (VLMs) to incorporate
multi-view observations, but two gaps remain. First, the \emph{visual-domain gap} limits transfer from simulation to
roadside surveillance. Methods such as VLMLight are developed primarily with
simulator-native renderings, which offer controllable traffic scenes but lack realistic
variation in illumination, weather, camera perspective, and visibility. Training visual
control therefore requires observations that combine roadside appearance diversity with
the traffic-state consistency needed for closed-loop simulation.

Second, the \emph{perception-to-decision gap} reflects a separation between visual
understanding and control optimization. Existing visual TSC pipelines commonly use
images to construct intermediate traffic representations or to assist a separate
controller~\cite{wang2025vlmlight,yao2025incorporating,pang2026reasonlight}.
The visual-to-action mapping thus remains split across perception and decision stages.
Bridging this gap calls for a unified policy that retains visual evidence while grounding
traffic demand in the movements served by each candidate phase. Reinforcement learning
can then align this visual-to-phase reasoning with the consequences of successive
decisions, optimizing the policy for long-term traffic outcomes.

To address these gaps, we jointly introduce SynTra-V for semantics-preserving Sim2Real
data generation and Traffic-VL for reinforced end-to-end visual-to-phase control.
We formulate this control task as a vision-language decision problem: visual observations
describe where traffic demand occurs, while language specifies which movements each
candidate phase serves. Jointly reasoning over these inputs connects visuospatial
traffic understanding to symbolic control semantics.

SynTra-V is a semantics-preserving Sim2Real data engine that transforms physics-based
traffic simulation into photorealistic multi-view surveillance imagery. For these images
to support control, appearance must vary without changing the traffic configuration
that determines the appropriate phase. The engine combines 2D-to-3D scene construction,
annotation-aligned rendering, structure-conditioned stylization, and automatic semantic
auditing. Structural references constrain vehicle layout and lane geometry during
generation, while audits check vehicle placement, movements, and lane semantics against
the simulator state. Across diverse topologies and sensing conditions, the resulting
observations support both visual-domain adaptation and realistic closed-loop evaluation
without requiring large-scale access to private roadside camera data.

Built on this data engine, Traffic-VL is an 8B vision-language model that directly
maps synchronized multi-view images and natural-language phase descriptions to phase
decisions without an external state-extraction stage (Fig.~\ref{fig:teaser}).
Each description identifies the movements served by
a permitted phase; the model locates their corresponding lanes in the images and compares
their observed demand. Describing phases by their movement semantics allows the same
controller to accommodate standard and irregular signal plans without topology-specific
output heads.

To learn this mapping, we extend Traffic-R1's two-stage reinforced reasoning framework
to visual control. Offline human-informed RL rewards expert-aligned phase decisions on
audited SynTra-V imagery, adapting both the vision encoder and reasoning model. The
online stage extends Stepwise Trajectory Policy Optimization to visual control (STPO-V),
using deterministic Blender renderings for efficient, temporally consistent rollouts.
We freeze the vision encoder to prevent updates on native simulator imagery from
overwriting the surveillance-domain adaptation acquired offline. The remaining policy
parameters are optimized for long-horizon traffic outcomes, separating visual-domain
adaptation from closed-loop policy optimization.

This journal extension inherits the two-stage RL framework and asynchronous
cross-intersection communication of Traffic-R1~\cite{zou2025trafficr1}. It replaces the
structured-text observation interface with direct multi-view perception, extends the
fixed four-phase template to natural-language descriptions of topology-dependent phase
sets, and introduces SynTra-V to support visual learning and evaluation.
New evaluations cover degraded sensing, zero-shot phase decisions on unseen
intersections in OmniTraffic and real surveillance images in Real-1296, and camera-based
field operation.
Table~\ref{tab:evolution} summarizes the changes.

Operating directly on multi-view images, Traffic-VL matches the privileged
structured-state Traffic-R1 across five closed-loop benchmarks while outperforming
image-based controllers and general-purpose VLMs. On real surveillance images, it
achieves 82.1\% phase-selection accuracy compared with 75.0\% for Gemini 3 Pro. In a
three-week field evaluation across ten interconnected intersections serving over 55,000
drivers daily, Traffic-VL reduces average queue length by 10.4\% relative to historical manual
operation at the same intersections. Our main contributions are:

\begin{itemize}

\item \textbf{SynTra-V, a semantics-preserving Sim2Real data engine.}
We combine structure-conditioned generation and control-critical semantic auditing to
produce diverse surveillance-style imagery for visual adaptation and photorealistic
closed-loop evaluation.

\item \textbf{Traffic-VL, an end-to-end vision-language TSC framework.}
We jointly ground multi-view traffic demand and natural-language phase semantics, enabling
direct visual decisions across heterogeneous road layouts and signal plans.

\item \textbf{Two-stage reinforced visual reasoning.}
We combine offline visual adaptation with online policy optimization using a frozen
vision encoder. Evaluation demonstrates structured-state-level control from raw images,
robustness under emergency traffic and sensing degradation, and transfer to unseen
topologies, real surveillance images, and constrained field operation.

\end{itemize}

\begin{table*}[t]
  \centering
  \caption{Comparison of Traffic-R1 and Traffic-VL across observation, reasoning, training, and deployment.}
  \label{tab:evolution}
  \small
  \setlength{\tabcolsep}{0pt}
  \renewcommand{\arraystretch}{1.16}
  \begin{tabular}{@{}L{0.12\textwidth}@{\hspace{0.8em}$|$\hspace{0.8em}}L{0.33\textwidth}@{\hspace{0.8em}$|$\hspace{0.8em}}L{0.45\textwidth}@{}}
    \toprule
    \textbf{Aspect} &
    \multicolumn{1}{c}{\textbf{Conference Version (Traffic-R1)}} &
    \multicolumn{1}{c}{\textbf{Journal Version (Traffic-VL)}} \\
    \midrule
    \textbf{Perception} &
    Structured text states from upstream perception &
    \textbf{Raw images} with \textbf{natural-language phase descriptions} \\
    \midrule
    \textbf{Reasoning} &
    LLM reasoning over pre-extracted states &
    Joint \textbf{visual and language reasoning} for phase selection \\
    \midrule
    \textbf{Phase Schema} &
    Fixed four-phase lane template &
    Language descriptions for \textbf{variable and complex phase sets} \\
    \midrule
    \textbf{Training Data} &
    Fixed textual symbolic traffic states &
    Photorealistic \textbf{imagery across topologies and environments} \\
    \midrule
    \textbf{Deployment} &
    Dedicated V2X detectors and CV pipelines &
    \textbf{Roadside-camera views} with natural-language phase descriptions \\
    \midrule
    \textbf{Generalization} &
    Standard intersections with fixed topology &
    Generalization across \textbf{topological and environmental variation} \\
    \bottomrule
  \end{tabular}
\end{table*}

\section{Related Work}

\subsection{RL-based Traffic Signal Control}

RL-based TSC learns signal policies from traffic observations and operational rewards~\cite{li2016traffic,wei2021survey}.
IntelliLight~\cite{wei2018intellilight} formulates intersection control through
traffic-specific states, actions, and rewards, while PressLight~\cite{wei2019presslight}
incorporates the max-pressure principle into policy learning. For network-wide control,
MPLight~\cite{chen2020mplight} scales decentralized pressure-based policies to large road
networks, and CoLight~\cite{wei2019colight} uses graph attention to capture interactions
among neighboring intersections. Together, these approaches establish effective mechanisms
for adapting signal decisions and coordinating traffic flow. Network-wide cooperation
is also studied through decentralized multi-agent RL~\cite{chu2020multiagent} and
graph-based cooperation~\cite{shang2025graph}.

Generalization also depends on how intersections are represented.
MetaLight~\cite{zang2020metalight} uses meta-learning to accelerate adaptation to new
traffic scenarios.
FRAP~\cite{zheng2019frap} models competition between phases and exploits symmetry in
traffic configurations. AttendLight~\cite{oroojlooy2020attendlight} uses attention to
accommodate varying numbers of roads, lanes, and phases, while
Advanced-CoLight~\cite{zhang2022advanced} enriches the state with pressure and running
vehicle demand. These advances improve state representations and policy transfer,
but their policies operate on predefined traffic measurements. Their transfer capability
therefore depends on the state and phase schemas supplied to the policy, even when
attention accommodates variable intersection sizes. Our focus extends this line of work
to learning control directly from multi-view visual evidence and
natural-language phase descriptions.

\subsection{Foundation Models for Traffic Signal Control}

Vision-language pretraining~\cite{radford2021clip,li2023blip2} and visual instruction
tuning~\cite{liu2023llava} support language-conditioned visual understanding, while
chain-of-thought prompting~\cite{wei2022cot} encourages intermediate reasoning steps.
PaLM-E~\cite{driess2023palme} and RT-2~\cite{brohan2023rt2} connect vision and language
to embodied planning and control; DriveGPT4~\cite{xu2024drivegpt4} and
DriveVLM~\cite{tian2024drivevlm} explore these capabilities in autonomous driving.
Within traffic signal control, foundation models introduce semantic reasoning into
phase selection.
LLMLight~\cite{lai2025llmlight} formulates phase selection as reasoning over textual
traffic observations and introduces LightGPT as a task-specialized backbone.
CoLLMLight~\cite{yuan2026collmlight} coordinates intersections through asynchronous
cooperative reasoning and refines its model with rewards balancing network performance
and reasoning cost. Our conference work
Traffic-R1~\cite{zou2025trafficr1} combines offline human-informed reinforcement learning,
online trajectory optimization, and asynchronous communication to improve generalization
and network-level coordination. These methods establish a foundation for reasoning-based
control, with traffic information supplied through structured or textual state interfaces.

Visual approaches expand the available evidence to include road appearance and
observable events. Yao et al.~\cite{yao2025incorporating} extract traffic states from
images for an LLM-based phase-selection agent. VLMLight~\cite{wang2025vlmlight}
introduces multi-view simulation and a meta-controller that selects between an RL policy
and a structured LLM reasoning branch. ReasonLight~\cite{pang2026reasonlight} combines
camera observations, sensor-derived descriptions, and an RL-proposed phase to refine
decisions for unseen events. These systems connect visual perception to control through
intermediate representations or interactions among specialized components.
Traffic-VL instead learns a unified visual-to-phase policy: the same model grounds
multi-view traffic demand in candidate phase semantics and is refined through expert
feedback and closed-loop traffic outcomes.

\subsection{Synthetic Traffic Data and Sim2Real Transfer}

Real-world datasets such as Cityscapes~\cite{cordts2016cityscapes} and
BDD100K~\cite{yu2020bdd100k} support visual understanding of road scenes.
SYNTHIA~\cite{ros2016synthia} and Playing for Data~\cite{richter2016playing} provide
synthetic images with semantic annotations. Traffic simulation additionally provides
controllable dynamics and annotations for data synthesis.
SUMO~\cite{lopez2018sumo} models microscopic traffic, while
CARLA~\cite{dosovitskiy2017carla} renders urban scenes with configurable sensors and
environmental conditions. OmniTraffic~\cite{wang2026omnitraffic} further turns
reconstructed intersections into synchronized multi-view supervision for traffic reasoning.
However, transferring models trained on simulated observations to roadside cameras
requires bridging the visual-domain gap while preserving the underlying traffic state.
Domain randomization~\cite{tobin2017domain} and appearance randomization with pyramid
consistency~\cite{yue2019domain} address visual transfer by varying synthetic appearances.

Conditional generation offers tools for visual adaptation:
ControlNet~\cite{zhang2023controlnet} adds spatial conditioning, and
MagicDrive~\cite{gao2024magicdrive} uses road maps, 3D boxes, and camera poses to generate
street scenes with cross-view consistency. For closed-loop TSC, however, generated
images must remain aligned with the simulated traffic dynamics: a displaced vehicle or
altered lane can change the appropriate phase even when an image appears realistic.
SynTra-V therefore couples simulator-derived structural conditions with semantic
auditing of vehicle placement, movements, and lane semantics. This connection enables
diverse surveillance-style imagery to support both visual adaptation and closed-loop
control evaluation while retaining the simulator's traffic-state correspondence.

\section{Preliminaries}
\label{sec:preliminaries}

\subsection{Road Network and Signal Phases}

We consider traffic signal control at an intersection in an urban road network. An
intersection consists of multiple incoming approaches, each containing one or more lanes.
A traffic movement is a directed maneuver from an incoming lane toward an outgoing
approach, typically categorized as a left turn, through movement, or right turn.
A signal phase activates a set of mutually compatible traffic movements. Let
\(\mathcal{P}\) denote the candidate phase set of an intersection and
\(\mathcal{M}(p)\) the movements released by phase \(p\in\mathcal{P}\).
For the four-approach example in Fig.~\ref{fig:phase-grounding}, Phase 1 serves the through
and right-turn movements of Views 1 and 3, and Phase 2 serves their left turns.
Phases 3 and 4 provide the corresponding movements for Views 2 and 4.
Both \(\mathcal{P}\) and \(\mathcal{M}(p)\) are determined by the intersection geometry, lane
configuration, and signal plan. They may therefore differ across standard four-leg,
T-shaped, and other irregular intersections.

\subsection{Synchronous Multi-View Observation}

At decision time \(t\), the controller receives an ordered tuple of synchronized roadside
images,
\begin{equation}
o_t=\left(I_t^{v_1},\ldots,I_t^{v_K}\right),
\end{equation}
where \(\mathcal{V}=\{v_1,\ldots,v_K\}\) denotes the \(K\) camera viewpoints associated
with the intersection. The viewpoints follow a fixed order and are aligned with the
incoming approaches. Each image may contain sensing imperfections encountered in roadside
surveillance, including occlusion, illumination variation, visual noise, and perspective
distortion.

We adopt a single-step observation protocol: the controller receives the current
synchronized images at each decision step, without an explicit history of previous frames.
Temporal effects are nevertheless reflected through repeated interaction with the evolving
traffic environment.

\subsection{Problem Formulation}

At each decision step, the controller selects a phase \(p_t\in\mathcal{P}\) from the
candidate set according to the current observation and available operational context:
\begin{equation}
p_t=\pi_{\theta}\left(o_t,\mathcal{P},c_t\right),
\end{equation}
where \(c_t\) denotes the coordination messages received from neighboring intersections
(Section~\ref{sec:coordination}). Timing and safety constraints are enforced by the
environment or the signal controller rather than inferred by the policy. Under the
traffic dynamics, the selected phase affects the subsequent state and observation. The
control objective is to maximize the expected cumulative reward,
\begin{equation}
\max_{\theta}\;
\mathbb{E}_{\pi_{\theta}}
\left[
\sum_{t=1}^{T}\gamma^{t-1}r(s_t,p_t)
\right],
\end{equation}
where \(s_t\) is the underlying traffic state, \(r(s_t,p_t)\) measures control quality,
and \(\gamma\) is the discount factor.

The policy produces a phase recommendation rather than directly actuating the signal.
During physical deployment, each recommendation is executed through the existing signal
controller under its approved timing and safety constraints.

\section{Methodology}
\subsection{Framework Overview}\label{sec:framework-overview}

\begin{figure*}[t]
  \centering
  \includegraphics[width=0.85\textwidth]{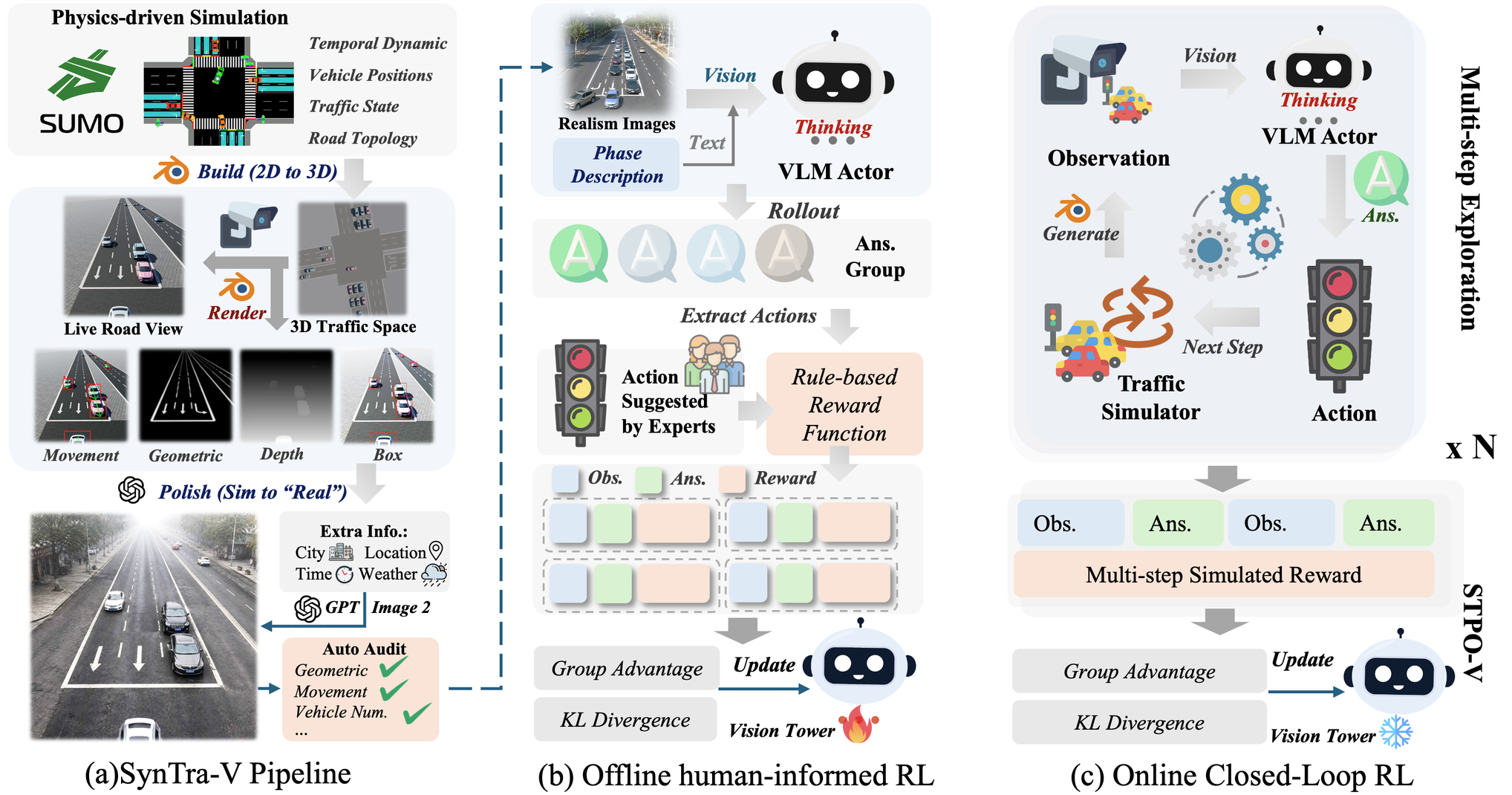}
  \caption{The Traffic-VL framework combines (a) SynTra-V data generation, (b) offline human-informed RL, and (c) online closed-loop policy optimization.}
  \label{fig:framework}
\end{figure*}

Fig.~\ref{fig:framework} summarizes the framework. SynTra-V (Section~\ref{sec:syntra})
converts traffic simulation into semantically audited, photorealistic multi-view
observations (Fig.~\ref{fig:framework}(a)). Traffic-VL (Section~\ref{sec:controller})
maps these observations and natural-language phase descriptions to phase recommendations.
It is trained in two stages (Section~\ref{sec:training}): offline human-informed RL on
SynTra-V images and expert labels (Fig.~\ref{fig:framework}(b)), followed by online
STPO-V in a closed simulator loop (Fig.~\ref{fig:framework}(c)). We first motivate this
visual-native formulation.

\subsection{Design Rationale for Visual-Native Control}
\label{sec:design-rationale}

Let \(T\) denote a fixed external perception pipeline that maps the multi-view
observation \(o_t\) to an estimated structured state \(\tilde{s}_t=T(o_t)\).
A controller operating on this representation selects
\begin{equation}
p_t
=
\pi_{\mathrm{str}}
\left(
T(o_t),\mathcal{P},c_t
\right).
\end{equation}

Any decision rule based on \(T(o_t)\) is also a function of \(o_t\).
However, if \(T\) discards evidence relevant to phase selection, the downstream
controller cannot recover it from the structured state. This motivates conditioning
Traffic-VL directly on the multi-view observation and candidate phase semantics,
keeping visual evidence available throughout the decision process.
Appendix~\ref{app:theory} formalizes this information-ceiling argument and the conditions
under which a structured representation preserves decision-relevant information.

Three design principles follow. \textbf{Evidence preservation} keeps visual tokens
available during phase reasoning. \textbf{Modality-native alignment} jointly grounds
visual traffic demand and symbolic phase semantics within a unified model.
\textbf{Semantics-preserving synthesis} varies appearance while maintaining
control-critical geometry, vehicle movements, and lane semantics.
These principles guide the controller, data engine, and two-stage training procedure.

\subsection{SynTra-V: Semantics-Preserving Sim2Real Data Engine}
\label{sec:syntra}

Real roadside camera data are difficult to acquire and annotate at scale because of
privacy, regulatory, and deployment constraints. TSC research therefore relies heavily on
traffic simulation, which provides closed-loop dynamics and exact traffic annotations at
low cost. However, symbolic simulator states omit perception errors altogether, while
simulator-native renderings rarely reproduce the appearance variation and sensing
imperfections of roadside surveillance. Models trained only under these idealized visual
conditions may therefore transfer poorly to physical intersections.

SynTra-V bridges this visual-domain gap by transforming dynamic traffic simulation into
photorealistic multi-view surveillance imagery while preserving the traffic semantics
required for phase selection. It retains the controllable dynamics and annotations of
simulation, but introduces realistic variations in topology, roadside appearance, weather,
illumination, camera perspective, and visibility. The resulting observations support both
visual-domain adaptation during offline training and photorealistic closed-loop evaluation.

\subsubsection{Simulation, Topology, and Scene Construction}

For scene construction, SynTra-V uses SUMO~\cite{lopez2018sumo}, which provides dynamic vehicle
positions, trajectories, traffic demand, signal states, and road topology. We construct
standard four-leg intersections together with irregular layouts, including T-shaped and
five-leg junctions. Each simulator state is converted into a three-dimensional traffic
scene containing road geometry, lane markings, vehicles, and roadside context.

Virtual cameras are placed at surveillance-style positions to render synchronized
multi-view observations. The underlying simulator state supplies exact vehicle positions,
counts, queues, movements, and lane assignments for annotation and auditing. Signal states
are available to the simulator but are not rendered as visible cues, preventing the model
from selecting phases by reading the current signal appearance rather than the traffic
demand.

For each topology, a rule-guided procedure constructs the candidate phase set from its
compatible traffic movements. Natural-language phase descriptions are then generated under
the schema introduced in Section~\ref{sec:phase-schema} and verified by traffic-domain
experts. This process exposes Traffic-VL to diverse road structures, lane configurations,
and signal plans without requiring topology-specific output heads.

\subsubsection{Structure-Conditioned Photorealistic Generation}

Although Blender rendering preserves traffic geometry, its native output lacks
the visual variability of roadside surveillance. SynTra-V therefore applies the
conditional image-to-image generator \texttt{gpt-image-2} under controllable contextual
conditions, including city style, location, time, weather, illumination, and camera
appearance.

As shown in Fig.~\ref{fig:framework}(a), the renderer produces a source view together with
four pixel-aligned auxiliary references: \emph{Box}, \emph{Movement},
\emph{Geometric}, and \emph{Depth}. The Box and Movement references are combined into a
vehicle control image \(C_{\mathrm{veh}}\), which constrains vehicle count, position, lane
occupancy, and heading. The Geometric reference is used to construct the lane control image
\(C_{\mathrm{lane}}\), preserving road geometry, lane markings, and turn arrows, including
regions occluded by vehicles. The Depth reference is retained for semantic auditing rather
than used as a generation condition.

\begin{figure}[t]
  \centering
  \includegraphics[width=\linewidth]{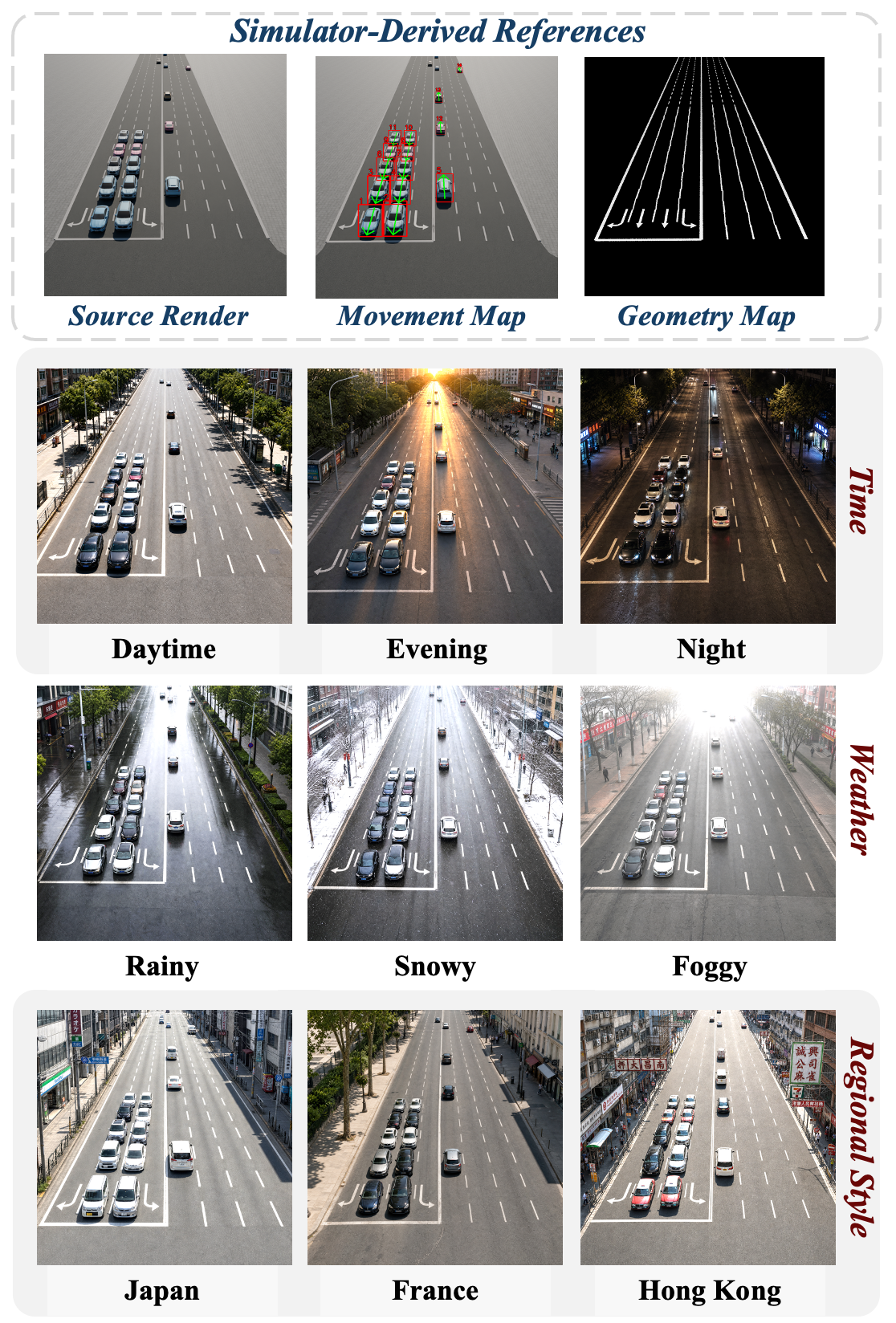}
  \caption{SynTra-V imagery across city styles and weather conditions.}
  \label{fig:syntra-examples}
\end{figure}

These references constrain the structure that determines the phase decision, leaving
appearance free to vary. Fig.~\ref{fig:syntra-examples} illustrates the appearance diversity produced from a shared simulator state. By varying scene attributes while retaining simulator-derived structural references, SynTra-V expands the same traffic configuration across different times, weather conditions, and regional road styles. These variations expose the vision encoder to surveillance-domain appearance shifts without requiring additional real-world annotations.

\subsubsection{Automatic Semantic Audit}

For each generated candidate, SynTra-V executes four independent Gemini 3 Pro audit calls.
Each call uses one of the pixel-aligned Blender references together with the generated image
and a task-specific auxiliary result: Box with a YOLO11n detection overlay, Geometric with a
morphological lane-and-arrow overlay, Depth with a monocular depth estimate, and Movement with
crop-level heading predictions. The four calls evaluate complementary dimensions:

\begin{itemize}
  \setlength{\itemsep}{0pt}
  \setlength{\topsep}{2pt}
  \item \textit{Box consistency:} verifies vehicle count and spatial placement.
  \item \textit{Geometric consistency:} verifies lane markings, turn arrows, road geometry,
  and vehicle-to-lane alignment.
  \item \textit{Depth consistency:} verifies near-to-far ordering and occlusion
  relationships.
  \item \textit{Movement consistency:} verifies vehicle headings and rejects reversed or
  physically implausible movements.
\end{itemize}

A candidate is accepted only if all four independent audit calls return \texttt{PASS};
otherwise, it is discarded
and resampled. Accepted images form the photorealistic observations used for offline visual
adaptation. The same generation and auditing protocol is used to construct observations
during photorealistic closed-loop evaluation. Appendix~\ref{app:audit} reports the generation
inputs and prompt, task-specific audit inputs, four audit prompts, and acceptance rule.

\subsection{End-to-End Visual-to-Phase Controller}
\label{sec:controller}

\subsubsection{Controller Overview}

Traffic-VL implements a direct mapping from multi-view visual observations to phase-level recommendations. Given the current observation \(o_t\), candidate phase set \(\mathcal P\), natural-language phase descriptions, and operational context \(c_t\), the controller encodes the ordered camera views into visual tokens. A single Qwen3-VL model jointly processes these tokens with the phase descriptions and task instruction. We use the 8B model as the primary configuration.

During inference, Traffic-VL identifies traffic demand from the images, grounds the movements referenced by each candidate phase in their corresponding visual lanes, and compares the candidate phases before producing a structured response. The response contains an optional reasoning trace and one recommended phase identifier.

\subsubsection{Multi-View Visual Encoding and View Alignment}

Traffic-VL uses the synchronized multi-view observation \(o_t\) defined in
Section~\ref{sec:preliminaries}. In the standard four-approach setting, the input contains
\(K=4\) roadside images at \(512\times512\) resolution, ordered by incoming approach. For
other layouts, \(K\) follows the number of observed approaches.

The fixed view order establishes a common spatial reference between the images and phase
descriptions. Each image is processed by the Qwen3-VL vision encoder, and the resulting
tokens retain their view order when passed to the language model. This organization allows
the controller to associate visual evidence with a specific approach while remaining
applicable to different numbers and arrangements of approaches.

\subsubsection{Natural-Language Phase Schema and Lane-to-Phase Grounding}
\label{sec:phase-schema}

A phase \(p\) is defined by its set of mutually compatible movements \(\mathcal M(p)\). To communicate these action semantics across heterogeneous intersections, we represent each candidate phase \(p\in\mathcal P\) using a natural-language descriptor \(D(p)\).

\begin{figure}[t]
  \centering
  \includegraphics[width=\linewidth]{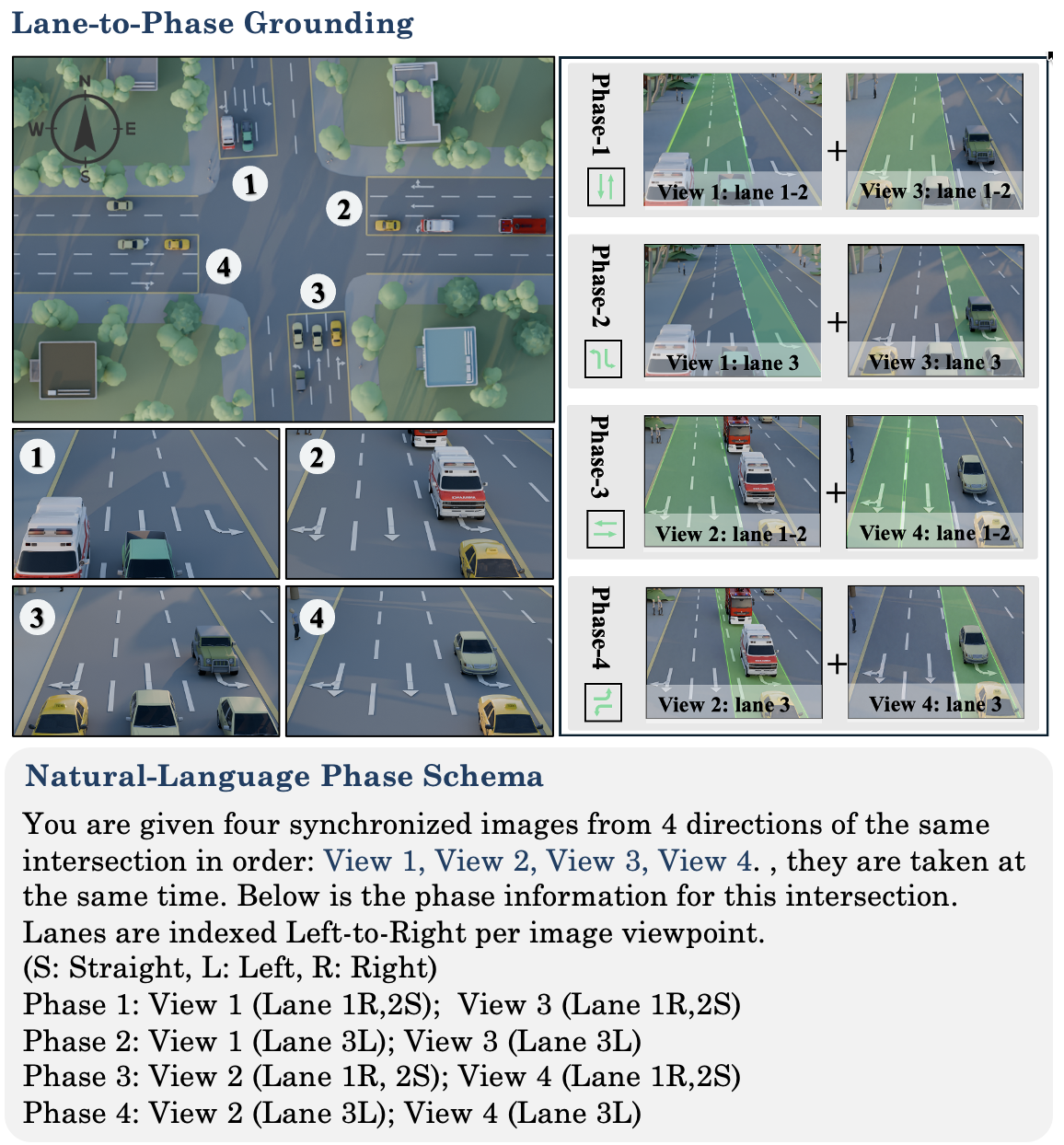}
  \caption{Natural-language phase descriptions and lane-to-phase grounding.}
  \label{fig:phase-grounding}
\end{figure}

As illustrated in Fig.~\ref{fig:phase-grounding}, each movement is addressed by
\begin{equation}
m=(v,\ell,d),
\end{equation}
where \(v\) identifies the source viewpoint and incoming approach, \(\ell\) is the lane index within that view, and \(d\in\{L,S,R\}\) denotes a left-turn, through, or right-turn movement. The descriptor \(D(p)\) lists the compatible movements released by phase \(p\) across the available viewpoints.

This representation separates phase semantics from the model architecture. Standard four-phase plans and nonstandard plans for T-shaped, five-leg, and other irregular intersections can be expressed using the same addressing convention. Introducing a new topology therefore changes the candidate phase descriptions rather than requiring a topology-specific output head.

The descriptor specifies the movements released by a phase but does not encode their current demand. Traffic-VL must locate each referenced movement in the corresponding image, infer its demand from visual evidence, and compare the candidate phases according to their movement composition. We refer to this process as \emph{lane-to-phase semantic grounding}. It dynamically connects symbolic phase semantics to visual lane regions instead of applying a fixed phase-to-lane mapping to pre-extracted traffic features.

\subsubsection{Reasoning and Valid Phase Selection}

The ordered visual tokens, phase descriptors, and task instruction are concatenated into a
single model context. Traffic-VL generates a structured response containing a reasoning
field and an answer field. The answer field must contain exactly one phase identifier from
the candidate set \(\mathcal P\). The reasoning trace supports visual comparison and phase
selection but is not parsed as part of the executed action.

The candidate set contains only conflict-free phases admitted by the intersection signal
plan. A response is valid if its answer field yields exactly one identifier in
\(\mathcal P\). During training, an unparsable response receives zero format and action
rewards. During simulation, an invalid response retains the previous valid phase. Yellow
and all-red clearance intervals are inserted by the environment and are not selected by
the model. 
The reward evaluates the validity and selected phase rather than supervising individual
reasoning tokens. Traffic-VL can therefore develop visual grounding and phase-comparison
strategies from outcome-level feedback without requiring manually authored reasoning traces
for every observation.

\subsection{Two-Stage Reinforcement Training}
\label{sec:training}

Figs.~\ref{fig:framework}(b) and~\ref{fig:framework}(c) summarize the two training
stages. Traffic-R1 previously established through controlled comparisons with supervised
fine-tuning that two-stage reinforcement learning improves zero-shot and safety-critical
generalization over structured traffic states. Traffic-VL extends this reinforced
reasoning paradigm to vision-language control.

Offline human-informed reinforcement learning uses semantically audited SynTra-V images
and expert phase labels to establish traffic-specific visual grounding and phase reasoning.
The vision encoder remains trainable in this stage. Online STPO-V then refines long-horizon
control through multi-step simulator interaction. To enable deterministic and scalable
rollouts, online training uses native Blender renderings and freezes the vision encoder,
confining policy updates to reasoning and phase selection.

\subsubsection{Offline Human-Informed Reinforcement Learning}

For an observation \(o\), phase descriptor set
\(\mathcal D=\{D(p)\mid p\in\mathcal P\}\), expert phase label \(a^\star\), and policy
\(\pi_\theta\), we sample a group of \(G\) structured responses
\(\{y_i\}_{i=1}^{G}\). Each response contains a reasoning trace and a parsed phase
\(\hat a(y_i)\). We assign the rule-based reward
\begin{equation}
R_i
=
\lambda_f R_{\mathrm{fmt}}(y_i)
+
\lambda_a
\mathbf 1\!\left[\hat a(y_i)=a^\star\right],
\label{eq:offline_reward}
\end{equation}
where \(R_{\mathrm{fmt}}(y_i)=1\) only if the response follows the prescribed output
format and contains one parseable phase identifier; otherwise it is zero. If no valid phase
can be parsed, the action reward is also zero.

We optimize the policy using Group Relative Policy Optimization (GRPO)~\cite{shao2024deepseekmath}. The group-normalized
advantage is
\begin{equation}
A_i
=
\frac{R_i-\mu_G}
{\sigma_G+\varepsilon_{\mathrm{std}}},
\end{equation}
where \(\mu_G\) and \(\sigma_G\) are the reward mean and standard deviation within the
sampled response group. The offline objective uses a clipped surrogate~\cite{schulman2017ppo}:
\begin{equation}
\begin{aligned}
\mathcal L_{\mathrm{offline}}
={}&
-\mathbb E\Bigg[
\frac{1}{G}
\sum_{i=1}^{G}
\frac{1}{|y_i|}
\sum_{u=1}^{|y_i|}
\min\Big(
\rho_{i,u}A_i,\\[-2pt]
&\hspace{1.25cm}
\operatorname{clip}_{\epsilon}\!(\rho_{i,u})A_i
\Big)\Bigg]
+\beta D_{\mathrm{KL}}
\!\left(\pi_\theta\,\Vert\,\pi_{\mathrm{ref}}\right),
\label{eq:offline_objective}
\end{aligned}
\end{equation}
where
\(\operatorname{clip}_{\epsilon}(z)
=\operatorname{clip}(z,1-\epsilon,1+\epsilon)\), and
\begin{equation}
\rho_{i,u}
=
\frac{
\pi_\theta
(y_{i,u}\mid o,\mathcal D,y_{i,<u})
}{
\pi_{\mathrm{old}}
(y_{i,u}\mid o,\mathcal D,y_{i,<u})
}.
\end{equation}

The expert annotations provide preferred phases rather than unique token-level reasoning
traces. The model therefore generates its own reasoning process and is optimized according
to output validity and agreement with the expert decision. Both the language model and
vision encoder are trainable during this stage, allowing the visual representation to
adapt to traffic-specific geometry, camera perspective, occlusion, weather, illumination,
and image degradation before closed-loop optimization.

\subsubsection{Online Closed-Loop Reinforcement Learning via STPO-V}

The online stage places Traffic-VL in a closed loop with CityFlow~\cite{zhang2019cityflow} traffic dynamics and
deterministic multi-view rendering. Applying stochastic photorealistic stylization at every
decision step would introduce appearance inconsistency across a trajectory and increase the
variance of multi-step credit assignment. Moreover, SynTra-V auditing has variable
generation latency, which limits parallel rollout throughput. Online STPO-V therefore uses
multi-view images rendered directly by Blender from the current simulator state. This
provides temporally consistent observations, predictable rendering cost, and scalable
parallel interaction.

At decision step \(t\), Traffic-VL receives observation \(o_t\), generates a structured
response \(y_t\), and selects phase \(a_t\). The simulator then transitions to state
\(s_{t+1}\). A trajectory of \(T\) control steps is
\begin{equation}
\tau
=
\left\{
(s_t,o_t,y_t,a_t)
\right\}_{t=1}^{T}.
\end{equation}

Let \(q_{t,l}\) and \(w_{t,l}\) denote the queue length and accumulated waiting time on
lane \(l\) at step \(t\). We define
\begin{equation}
\bar q_t
=
\frac{1}{|\mathcal L|}
\sum_{l\in\mathcal L}q_{t,l},
\qquad
\bar w_t
=
\frac{1}{|\mathcal L|}
\sum_{l\in\mathcal L}w_{t,l},
\end{equation}
and the operational cost
\begin{equation}
c(s_t)
=
\omega_q\bar q_t
+
\omega_w\bar w_t.
\label{eq:operational_cost}
\end{equation}
Each cost is measured after the selected phase has been executed, and the resulting
state also provides the next observation. The trajectory return is
\begin{equation}
R_{\mathrm{traj}}(\tau)
=
-\sum_{t=1}^{T}c(s_{t+1}).
\label{eq:trajectory_return}
\end{equation}

Following the Stepwise Trajectory Policy Optimization rule of Traffic-R1, the trajectory
return is broadcast to its \(T\) decision responses with length normalization:
\begin{equation}
r_t
=
\mathcal C_t\!\left(R_{\mathrm{traj}}\right)
=
\frac{R_{\mathrm{traj}}(\tau)}{T}.
\label{eq:stpo_credit}
\end{equation}
For \(G_{\mathrm{on}}\) rollouts generated from the same initial traffic condition, the
stepwise group advantage is
\begin{equation}
A_t^{(j)}
=
\frac{
r_t^{(j)}-\mu_t
}{
\sigma_t+\varepsilon_{\mathrm{std}}
},
\end{equation}
where \(\mu_t\) and \(\sigma_t\) are computed across the rollout group.

Here, \emph{stepwise} means that every response along the trajectory enters the policy
objective under the shared long-horizon outcome; it does not introduce a separate
temporal-difference target within the trajectory.
STPO-V optimizes the same clipped, KL-regularized policy objective as
Eq.~\ref{eq:offline_objective}, replacing the offline expert-derived advantage \(A_i\)
with the trajectory-derived advantage \(A_t^{(j)}\). Each structured response in the
trajectory is therefore optimized against the long-horizon outcome of its rollout rather
than only its agreement with a local expert decision.

The vision encoder is frozen throughout online optimization, while the language-model
reasoning and phase-selection parameters remain trainable. This prevents deterministic
Blender imagery from overwriting the surveillance-style visual representation learned from
SynTra-V. Online learning can then focus on multi-step phase comparison and adaptation to
closed-loop traffic dynamics while preserving the end-to-end mapping from images to phase
recommendations.

\subsection{Asynchronous Cross-Intersection Coordination}
\label{sec:coordination}

Traffic-VL retains the asynchronous coordination mechanism of Traffic-R1 for
multi-intersection control. Let the road network be
\(G=(\mathcal I,\mathcal E)\). Intersections are partitioned into two groups,
\(\mathcal I_1\) and \(\mathcal I_2\), according to positional parity.

\begin{figure}[t]
  \centering
  \includegraphics[width=\linewidth]{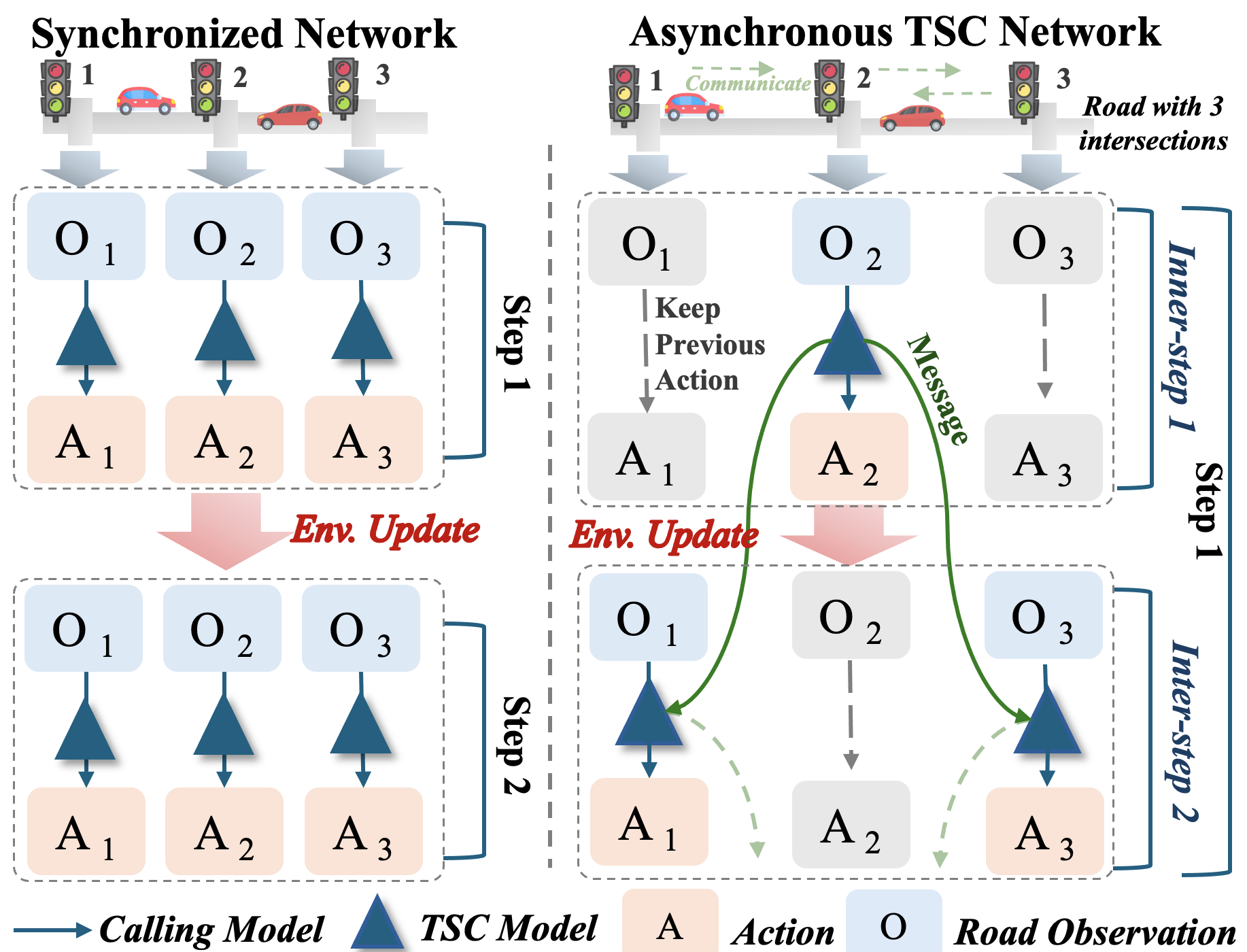}
  \caption{Synchronous control updates all intersections together; asynchronous control alternates two groups with half-step message exchange.}
  \label{fig:async_network}
\end{figure}

As illustrated in Fig.~\ref{fig:async_network}, each control step consists of two inner
steps. Intersections in \(\mathcal I_1\) act first using their local visual observations
and messages received from \(\mathcal I_2\) during the preceding inner step. Their outgoing
messages are then delivered to eligible intersections in \(\mathcal I_2\), which act in
the second inner step. An inactive intersection retains its previous valid phase until its
group is activated.

Communication is restricted to connected intersections within 2\,km, and each message
summarizes the directional traffic expected to reach the receiver.
Appendix~\ref{app:async_algorithm} details the message format, lifecycle, and half-step
execution procedure.

\section{Experimental Setup}

We evaluate Traffic-VL as a vision-language controller that acts directly on synchronized multi-view images. Our experiments address five research questions. \textbf{RQ1}: Can Traffic-VL preserve the control quality of privileged structured-state
controllers while operating directly on raw multi-view images? \textbf{RQ2}: Can it handle safety-critical and abnormal traffic conditions, including emergency prioritization and degraded sensing? \textbf{RQ3}: Does it generalize zero-shot to unseen intersection topologies and real surveillance frames? \textbf{RQ4}: What is the contribution of each component? \textbf{RQ5}: Can Traffic-VL meet the real-time control deadline on local server-class hardware, and how does model scale affect latency and accuracy? We first describe the experimental setup and then address these questions in turn.

\subsection{Implementation Details}

Traffic-VL uses Qwen3-VL 8B as its primary backbone; Qwen3-VL 4B is included only in the
scaling study. As detailed in Section~\ref{sec:training}, training consists of offline
human-informed GRPO followed by online STPO-V. We reuse the phase annotations collected for
Traffic-R1 from 11 traffic experts and use SynTra-V to construct approximately 5,000
photorealistic multi-view training samples spanning diverse city styles and traffic
densities.

Both stages use a constant learning rate of $1\times10^{-6}$, a global batch size of 256, a
clipping range of 0.2--0.3, and a KL coefficient of 0.01. Prompts and responses are capped
at 2048 tokens. Training is conducted on four NVIDIA A800 GPUs with 80\,GB memory each.

During offline GRPO, we train for 45 epochs using AdamW~\cite{loshchilov2019adamw} with
$(\beta_1,\beta_2)=(0.9,0.999)$, weight decay 0.01, and no warmup. For each state, we sample
five responses with temperature 1.0 and top-$p$ 1.0; validation uses one response with
temperature 0.6 and top-$p$ 0.95. The complete actor, including the vision tower, is
optimized end to end with BF16 FSDP.

Online STPO-V is performed in CityFlow on the same $4\times4$ simulated network as Traffic-R1, using
trajectory length $T=5$ and $G_{\mathrm{on}}=5$ rollouts per initial condition. The vision
encoder is frozen to preserve the visual representation learned offline, while all remaining
parameters are fully optimized. Neither stage uses LoRA or any other parameter-efficient
adaptation method.

\subsection{Environments and Datasets}

We evaluate closed-loop control in CityFlow on the five public Jinan and Hangzhou benchmarks used in
Traffic-R1: Jinan1, Jinan2, Jinan3, Hangzhou1, and Hangzhou2. Unlike online STPO-V
training, which uses deterministic Blender renders, these evaluations employ the complete
SynTra-V pipeline. At each decision step, the current simulator state is rendered into four
synchronized photorealistic roadside views and passed through the semantic audit before
being presented to the controller. The selected phase then updates the simulator, from
which the next visual observation is generated, forming a stepwise closed control loop.
All raw-image methods follow the same observation-generation and auditing protocol.
Although the evaluation is closed-loop, it is not constrained by wall-clock execution
time: the simulator advances only after a valid observation and control decision are
obtained. Image-generation latency therefore does not affect the simulated traffic dynamics
or the reported control metrics. Following Traffic-R1, each selected phase is activated for
15\,s, followed by a 3\,s yellow interval and a 2\,s all-red clearance interval.
Closed-loop results are averaged over five paired traffic-demand seeds. City style, weather,
illumination, and camera parameters remain fixed within each rollout and are resampled across
seeds; every controller is evaluated under the same seed-specific conditions.
Appendix~\ref{app:statistics} reports dispersion and paired equivalence tests for the two principal
controllers.

Zero-shot topology generalization is evaluated on twelve held-out OmniTraffic intersections~\cite{wang2026omnitraffic},
none of which appears during training. OmniTraffic retains its native simulator-rendered
observations. We further evaluate Sim2Real generalization on Real-1296, a fixed benchmark of
1,296 synchronized multi-view surveillance samples.
Real-1296 contains anonymized samples from 14 intersections that are disjoint from both the
training networks and the ten deployment intersections. The model, prompts, and evaluation
protocol are frozen before any benchmark result is inspected, and no sample is used for
model selection. Samples from the same camera are separated by at least 300\,s to limit
temporal duplication. The collection spans daytime and nighttime operation, rain, strong
occlusion, visual noise, and heterogeneous intersection layouts. License plates and faces
are removed or blurred before annotation. Three traffic-domain experts independently label
each sample with the preferred next phase, followed by adjudication when their initial
labels differ. Inter-expert agreement is 0.84 by Fleiss' $\kappa$. Multiple acceptable phases
are retained when two or more choices are operationally equivalent; these cases account for
approximately 15\% of the benchmark.

\subsection{Baselines}

We compare Traffic-VL with three groups of baselines. \textit{Structured-state controllers} access simulator-provided lane-level states. They include rule-based methods (FixedTime~\cite{koonce2008traffic} and MaxPressure~\cite{varaiya2013maxpressure}), RL methods (MPLight~\cite{chen2020mplight}, CoLight~\cite{wei2019colight}, Advanced-CoLight~\cite{zhang2022advanced}, and AttendLight~\cite{oroojlooy2020attendlight}), and language-based controllers that reason over structured textual observations (LLMLight~\cite{lai2025llmlight}, CoLLMLight~\cite{yuan2026collmlight}, Traffic-R1~\cite{zou2025trafficr1}, and the original VLMLight~\cite{wang2025vlmlight}).

\textit{Raw-image baselines} receive the same raw multi-view images as Traffic-VL. They include the unfine-tuned Qwen3-VL 8B base model~\cite{bai2025qwen3vl}, Gemini 3 Pro~\cite{deepmind2025gemini3pro}, and VLMLight (adapted). For VLMLight (adapted), we retain its original 32B scene-understanding VLM and 72B reasoning LLM while replacing its simulator-native views with the same SynTra-V observations used by Traffic-VL.

We further include Advanced-CoLight$\dagger$ as a full-shot perception-to-structure baseline. It applies a fine-tuned Grounding DINO detector~\cite{liu2024groundingdino} to multi-view images and uses manually specified, intersection-specific mappings to derive the lane-level queue and pressure states required by Advanced-CoLight. The mappings are calibrated separately for the Jinan and Hangzhou layouts, and the controller is trained on each target benchmark. This baseline exposes the deployment cost of conventional visual pipelines, which require detector integration, topology-specific state engineering, and target-environment training.

Following Traffic-R1, all learning-based baselines except Advanced-CoLight$\dagger$ are trained in a shared simulated environment and evaluated zero-shot on the test benchmarks, without benchmark-specific retraining. Since conventional RL controllers do not reliably transfer across topologies under this protocol, their full-shot results are additionally reported in Appendix~\ref{app:fullshot}. Traffic-VL is never trained on the evaluation benchmarks and acts at each unseen intersection without additional adaptation.

\subsection{Metrics}

We use metrics matched to each evaluation setting. For standard closed-loop control, we report Average Travel Time (ATT) and Average Waiting Time (AWT), in seconds, where lower is better. For zero-shot topology generalization on OmniTraffic, we report phase-selection accuracy. For real surveillance frames, we report expert-agreement accuracy, defined as the fraction of predicted phases matching traffic-expert annotations.
Following Traffic-R1, safety-critical evaluations use Emergency Action Accuracy (EAA) for local incidents, such as on-site accidents, and Average Emergency Travel Time (AETT) and Average Emergency Waiting Time (AEWT) for network-level incidents, such as emergency-vehicle routing. EAA measures the fraction of correct priority actions, while lower AETT and AEWT indicate better emergency response. We additionally report end-to-end decision latency to assess deployment efficiency.

In result tables, Depl. denotes direct camera-based deployment without topology-specific
structured-state engineering; bold and underlining mark the best and second-best values,
respectively. A dash in a metric column denotes an unevaluated result.

\section{Experiments}
\label{sec:experiments}

\subsection{RQ1: Overall Performance}

We compare closed-loop control under each method's native observation interface.
Structured-state baselines receive simulator-provided lane-level states, whereas Traffic-VL
operates directly on raw multi-view images. The central question is whether end-to-end
vision-language reasoning can preserve the control quality of a privileged structured-state
controller under a more realistic observation setting.

We include two bridge baselines. VLMLight (adapted) receives the same raw multi-view images
as Traffic-VL while retaining its original scene-understanding and reasoning pipeline. Advanced-CoLight$\dagger$
represents an engineered perception-to-structure pipeline: detections from a fine-tuned Grounding DINO model are
converted into queue and pressure states through manually specified, topology-dependent
mappings, after which the controller is trained in the full-shot setting. These baselines
separate direct visual reasoning from adapting an existing VLM controller or reconstructing
a conventional structured-state interface.

\begin{table*}[t]
  \centering
  \caption{Closed-loop traffic signal control on five public benchmarks (ATT/AWT in seconds; five paired traffic-demand seeds).}
  \label{tab:rq1}
  \small
  \setlength{\tabcolsep}{2.4pt}
  \begin{tabular}{@{}l l l c rr rr rr rr rr@{}}
    \toprule
    \multirow{2}{*}{\textbf{Method}} & \multirow{2}{*}{\textbf{Input}} &
    \multirow{2}{*}{\textbf{Setting}} & \multirow{2}{*}{\textbf{Depl.}} &
    \multicolumn{2}{c}{\textbf{Jinan1}} & \multicolumn{2}{c}{\textbf{Jinan2}} &
    \multicolumn{2}{c}{\textbf{Jinan3}} & \multicolumn{2}{c}{\textbf{Hangzhou1}} &
    \multicolumn{2}{c}{\textbf{Hangzhou2}} \\
    \cmidrule(l){5-6} \cmidrule(l){7-8} \cmidrule(l){9-10} \cmidrule(l){11-12} \cmidrule(l){13-14}
    & & & & ATT & AWT & ATT & AWT & ATT & AWT & ATT & AWT & ATT & AWT \\
    \midrule
    FixedTime        & Structured-Oracle        & Rule-based & \ding{55} & 453.41 & 51.32 & 370.34 & 35.15 & 384.53 & 36.95 & 497.54 & 36.41 & 408.53 & 53.94 \\
    MaxPressure      & Structured-Oracle        & Rule-based & \ding{55} & 274.34 & 32.04 & 246.35 & 22.56 & 245.66 & 24.31 & 289.55 & 21.52 & 349.53 & 67.52 \\
    MPLight          & Structured-Oracle        & Zero-shot  & \ding{55} & 455.34 & 72.45 & 471.14 & 78.03 & 427.37 & 64.91 & 491.32 & 64.05 & 425.42 & 69.85 \\
    CoLight          & Structured-Oracle        & Zero-shot  & \ding{55} & 472.44 & 91.09 & 450.41 & 78.59 & 498.84 & 89.94 & 494.61 & 72.18 & 435.32 & 81.11 \\
    Advanced-CoLight & Structured-Oracle        & Zero-shot  & \ding{55} & 347.31 & 56.54 & 345.78 & 35.96 & 342.56 & 37.55 & 485.32 & 54.11 & 523.19 & 72.56 \\
    AttendLight      & Structured-Oracle        & Zero-shot  & \ding{55} & 381.11 & 67.59 & 305.53 & 64.72 & 331.34 & 66.42 & 318.94 & 67.84 & 348.41 & 65.58 \\
    LLMLight         & Structured-Oracle (text) & Zero-shot  & \ding{55} & 274.47 & 33.66 & 286.53 & 28.66 & 271.11 & 28.27 & 299.31 & 25.53 & 331.38 & 51.79 \\
    CoLLMLight       & Structured-Oracle (text) & Zero-shot  & \ding{55} & 281.12 & 33.23 & 269.34 & 25.51 & 268.32 & 34.36 & 298.42 & 24.45 & 336.92 & 45.43 \\
    Traffic-R1       & Structured-Oracle (text) & Zero-shot  & \ding{55} & \underline{270.34} & \underline{27.95} & \textbf{239.53} & \underline{21.11} & \underline{238.03} & \textbf{23.17} & \underline{277.83} & \textbf{15.51} & \underline{324.11} & \underline{33.14} \\
    \midrule
    Qwen3-VL 8B (base)        & Raw-Image             & Zero-shot & \ding{51} & 372.32 & 47.31 & 331.56 & 43.22 & 325.81 & 42.08 & 386.09 & 49.19 & 449.24 & 80.37 \\
    Gemini 3 Pro              & Raw-Image             & Zero-shot & \ding{51} & 333.06 & 41.10 & 297.79 & 36.94 & 289.92 & 35.76 & 340.23 & 42.34 & 403.88 & 69.57 \\
    VLMLight (adapted)        & Raw-Image             & Zero-shot & \ding{51} & 300.53 & 36.04 & 263.47 & 31.53 & 266.11 & 32.42 & 305.76 & 37.23 & 357.55 & 61.02 \\
    Advanced-CoLight$\dagger$ & Raw-Image$\to$Struct  & Full-shot   & \ding{55} & 315.03 & 40.05 & 277.44 & 34.89 & 281.27 & 35.89 & 324.50 & 41.62 & 376.97 & 66.41 \\
    \textbf{Traffic-VL (ours)} & Raw-Image            & Zero-shot & \ding{51} & \textbf{269.71} & \textbf{27.82} & \underline{239.81} & \textbf{21.05} & \textbf{237.95} & \underline{23.31} & \textbf{277.56} & \underline{15.62} & \textbf{323.89} & \textbf{33.12} \\
    \bottomrule
  \end{tabular}
\end{table*}

Table~\ref{tab:rq1} shows that Traffic-VL matches the privileged structured-state reference,
Traffic-R1, across all five Jinan and Hangzhou benchmarks. Paired two one-sided tests (TOST)
with a pre-specified $\pm2\%$ relative equivalence margin support equivalence in both ATT and
AWT on every benchmark after Holm correction for multiple comparisons
(Appendix~\ref{app:statistics}). Traffic-VL therefore
preserves structured-state-level control quality while inferring traffic demand directly
from raw images.

The advantage is clearer against alternative raw-image controllers. On Jinan1, Traffic-VL
reduces ATT by approximately 19\% relative to Gemini 3 Pro and by 10\% relative to VLMLight
(adapted). It also outperforms Advanced-CoLight$\dagger$, whose engineered
perception-to-structure pipeline reaches 315.03\,s despite topology-specific state design and
full-shot training. Conventional RL controllers transfer less effectively under the
zero-shot protocol: Advanced-CoLight, even with structured inputs, reaches 347.31\,s ATT and
trails the rule-based MaxPressure baseline. Together, these results show that Traffic-VL
closes the gap to the privileged structured-state reference while acting directly on
camera images. When retrained on each target benchmark, conventional RL controllers become
competitive, and full-shot Advanced-CoLight attains the lowest ATT on three of the five
benchmarks (Appendix~\ref{app:fullshot}); Traffic-VL reaches comparable performance
without benchmark-specific training.

\subsection{RQ2: Safety-Critical and Abnormal Scenes}

\subsubsection{Emergency-Vehicle Prioritization}

Following Traffic-R1, we evaluate emergency-vehicle prioritization at both the local-intersection and three-intersection corridor levels, using EAA, AETT, and AEWT. The central difference is the observation interface. Traffic-R1 receives an emergency annotation in its structured text state, whereas Traffic-VL receives only synchronized raw images and must identify the emergency, interpret the traffic context, and select a phase directly from visual evidence. SynTra-V renders emergency vehicles and lane-blocking accidents into the multi-view observations. Text-based baselines receive oracle emergency annotations, while conventional RL baselines retain count-based states without vehicle-class information.

\begin{table}[t]
  \centering
  \caption{Emergency-vehicle prioritization: local EAA and corridor AETT/AEWT (s).}
  \label{tab:rq2}
  \small
  \setlength{\tabcolsep}{2.5pt}
  \begin{adjustbox}{max width=\linewidth}
  \begin{tabular}{@{}L{2.3cm} L{1.6cm} l c rrr@{}}
    \toprule
    \textbf{Method} & \textbf{Input} & \textbf{Scale} & \textbf{Depl.} &
    \textbf{EAA} & \textbf{AETT} & \textbf{AEWT} \\
    \midrule
    MaxPressure                & Structured-Oracle        & --       & \ding{55} & 0.20 & 287.94 & 21.87 \\
    Advanced-CoLight           & Structured-Oracle        & $\sim$1M & \ding{55} & 0.30 & 286.32 & 24.53 \\
    LLMLight (oracle emerg.)   & Structured-Oracle (text) & 7B       & \ding{55} & 0.42 & 273.55 & 15.21 \\
    Traffic-R1 (oracle emerg.) & Structured-Oracle (text) & 3B       & \ding{55} & \textbf{0.85} & \textbf{215.58} & \textbf{7.98} \\
    Advanced-CoLight$\dagger$  & Raw-Img$\to$Str.         & $\sim$1M & \ding{55} & 0.34 & 289.03  & 25.81  \\
    \midrule
    Gemini 3 Pro               & Raw-Image                & Undisclosed & \ding{51} & 0.71 & 232.06  & 13.15  \\
    VLMLight (adapted)         & Raw-Image                & 32B+72B  & \ding{51} & 0.82 & 226.10  & 11.22  \\
    \textbf{Traffic-VL (ours)} & Raw-Image       & 8B       & \ding{51} & \underline{0.83} & \underline{219.09} & \underline{8.77} \\
    \bottomrule
  \end{tabular}
  \end{adjustbox}
\end{table}

Table~\ref{tab:rq2} first confirms the limitation of count-based control: without vehicle-class information, conventional RL controllers cannot reliably prioritize emergency vehicles. Under the same oracle incident descriptions, Traffic-R1 substantially outperforms the SFT-based LLMLight. This gap further validates the reinforced reasoning paradigm of Traffic-R1, in which two-stage RL improves both traffic-semantic reasoning and generalization to safety-critical events beyond supervised demonstrations.

Despite receiving no incident description, Traffic-VL approaches the oracle-input Traffic-R1 on local emergency actions and achieves the strongest results among raw-image methods in corridor-level emergency response. These results show that direct vision-language reasoning can support both emergency recognition and coordinated control without privileged annotations or an external structured-state pipeline. Traffic-VL achieves this with a single 8B model, compared with VLMLight's 32B vision model and 72B reasoning model.

\subsubsection{Robustness under Degraded Perception}

We further evaluate robustness to degraded visual observations, a common failure mode in physical deployments. Occlusion, unstable illumination, and cropping can corrupt detector outputs; in a perception-to-structure pipeline, the resulting errors in lane counts and geometry are directly propagated to the downstream controller.

\begin{figure}[t]
  \centering
  \includegraphics[width=0.9\linewidth]{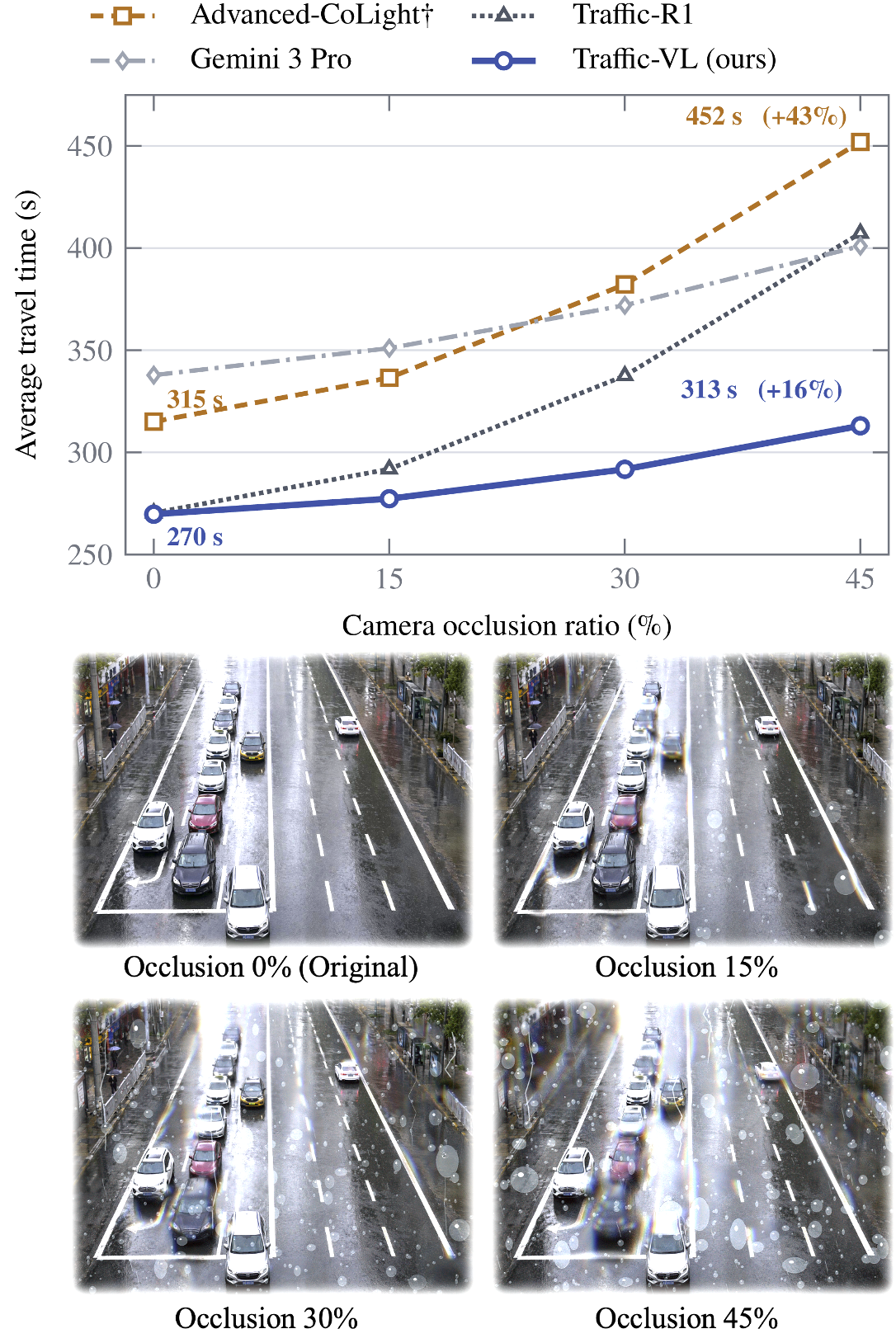}
  \caption{Closed-loop travel time (top) and example camera views (bottom) under increasing rain-induced occlusion on Jinan1.}
  \label{fig:rq3}
\end{figure}

Fig.~\ref{fig:rq3} reports closed-loop travel time on Jinan 1 as camera occlusion increases. Advanced-CoLight$\dagger$, our perception-to-structure bridge baseline, first applies a fine-tuned Grounding DINO detector and topology-specific mappings to convert images into the structured queue and pressure states required by its controller. Its travel time rises from 315.03\,s to 452\,s at 45\% occlusion, a 43\% increase. In contrast, Traffic-VL increases from 269.71\,s to 313\,s, or 16\%, under the same condition.

Traffic-R1 receives the structured text produced by its CV perception pipeline from the same occluded views. It matches Traffic-VL on clean observations, but as occlusion increases, the pipeline reports erroneous queue lengths and its travel time rises in the same pattern as Advanced-CoLight$\dagger$. Traffic-VL instead reasons over the partially occluded views directly and maintains substantially more stable control.

\subsection{RQ3: Zero-Shot Generalization}

\subsubsection{Phase Decision on Unseen Intersections}
\begin{figure*}[t]
  \centering
  \includegraphics[width=0.95\linewidth]{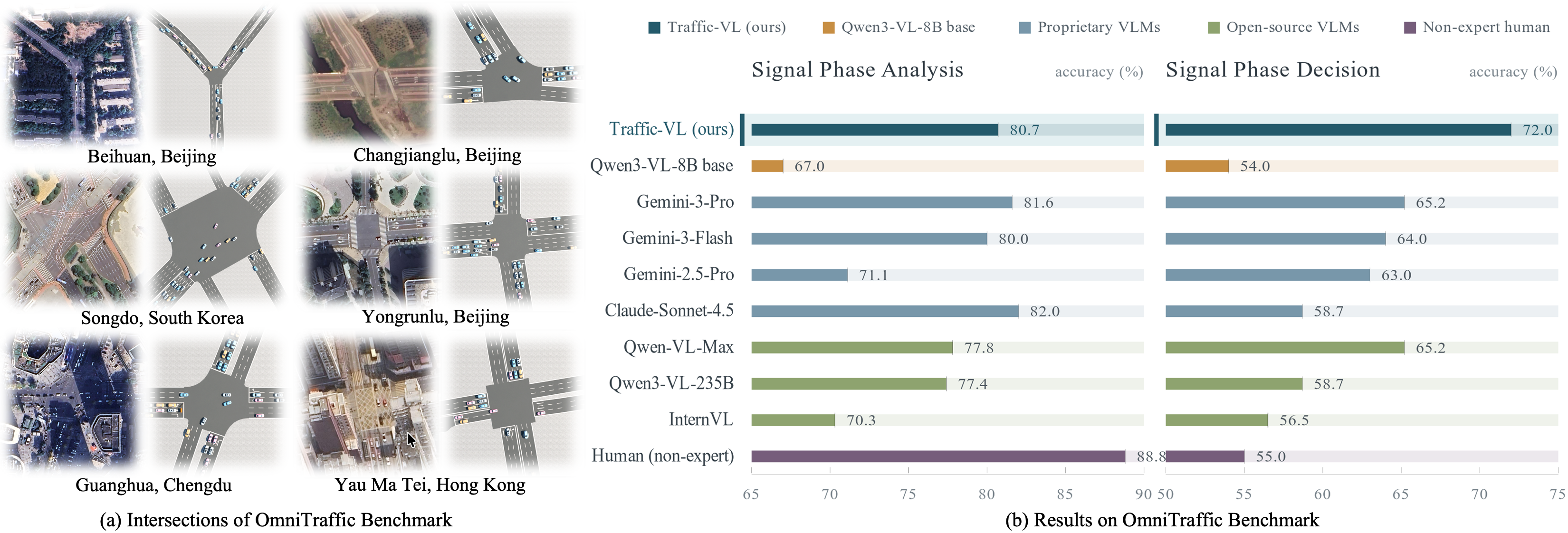}
  \caption{Zero-shot evaluation on OmniTraffic: (a) satellite views and 3D intersection reconstructions; (b) phase-analysis and phase-decision accuracy.}
  \label{fig:rq4}
\end{figure*}

A controller deployed at a new intersection must handle a topology and signal plan absent
from training. We evaluate this setting on OmniTraffic, which reconstructs twelve
real-world intersections from several cities in China, France, Hong Kong, and South Korea
as controllable 3D environments (Fig.~\ref{fig:rq4}(a)). For each environment, the
benchmark renders synchronized views of the four incoming approaches and provides the
definition of every candidate phase. We use its two decision-support tasks. Signal Phase
Analysis (Phase Ana.) asks which phase is most relevant to the current scene, such as the
phase whose movements are blocked by an accident or contain an emergency vehicle. Signal
Phase Decision (Phase Dec.) corresponds to the control task of Traffic-VL: the model
selects the next green phase, and the reference answer is fixed by a traffic-engineering
rule. None of the twelve intersections appears in training. We compare Traffic-VL with
the seven frontier multimodal models reported on the benchmark.

On Phase Dec., Traffic-VL reaches 72.0\%, about 7 points above the strongest general
models, Gemini 3 Pro~\cite{deepmind2025gemini3pro} and Qwen-VL-Max~\cite{bai2023qwenvl}
(65.2\%) (Fig.~\ref{fig:rq4}(b)). Gemini-3-Flash reaches 64.0\%, and the remaining models
fall between 56\% and 63\%. Traffic-VL outperforms Gemini 3 Pro at each of the twelve
intersections, with per-intersection accuracy from 64.3\% to 80.7\%. Absolute scores
remain moderate for all methods because each scene has a single rule-defined answer and
many scenes are near-ties decided by phase-transition logic; non-expert human annotators
reach 55.0\%. The same-backbone comparison isolates the effect of training: the
unfine-tuned Qwen3-VL-8B scores 54.0\%, 18 points below Traffic-VL.

Although Traffic-VL is never trained on Phase Ana., it reaches 80.7\% on this task, close
to Claude Sonnet 4.5~\cite{anthropic2025sonnet45} (82.0\%) and Gemini 3 Pro (81.6\%) and
well above its 8B base (67.0\%). Training for phase decisions therefore also improves the
recognition of phase-relevant scene events.

\subsubsection{From Simulation to Real Frames}

We finally evaluate Sim2Real generalization on the fixed Real-1296 benchmark in a zero-shot,
open-loop setting. Accuracy is the fraction of samples for which the predicted phase matches
at least one expert-approved phase, as reported in Table~\ref{tab:rq3}. Unlike OmniTraffic,
which assigns a single rule-defined optimal phase and excludes ambiguous scenes, Real-1296
permits multiple acceptable phases when they are operationally equivalent. Structured-state
methods, including Traffic-R1, are excluded because they cannot operate directly on raw
camera observations without intersection-specific perception engineering.
\begin{table}[t]
  \centering
  \caption{Zero-shot phase-selection accuracy on Real-1296, with paired-bootstrap 95\% confidence intervals.}
  \label{tab:rq3}
  \small
  \renewcommand{\arraystretch}{1.10}
  \setlength{\tabcolsep}{3pt}
  \begin{adjustbox}{max width=\linewidth}
  \begin{tabular}{@{}l l rr@{}}
    \toprule
    \textbf{Method} & \textbf{Scale} & \textbf{Acc. (\%)} & \textbf{95\% CI} \\
    \midrule
    Qwen3-VL (base)       & 8B        & 56.4 & 53.7--59.1 \\
    GPT-5.2               & Undisc.   & 70.0 & 67.5--72.5 \\
    VLMLight (adapted)    & 32B+72B   & 73.5 & 71.1--75.9 \\
    Gemini 3 Pro          & Undisc.   & \underline{75.0} & 72.7--77.3 \\
    Traffic-VL w/o SynTra-V & 8B     & 66.0 & 63.4--68.6 \\
    \addlinespace[1pt]
    \textbf{Traffic-VL (ours)} & \textbf{8B} & \textbf{82.1} & \textbf{80.0--84.1} \\
    \bottomrule
  \end{tabular}
  \end{adjustbox}
\end{table}

Traffic-VL achieves 82.1\% accuracy, compared with 75.0\% for Gemini 3 Pro and 73.5\% for
VLMLight (adapted). A paired bootstrap estimates a 7.1-point improvement over Gemini 3 Pro
(95\% CI: 4.4--9.8 points, $p<0.001$); the corresponding 95\% confidence interval for
Traffic-VL accuracy is 80.0--84.1\%. The result shows that traffic-specific visual grounding
transfers across real intersections and sensing conditions that are absent from training.

\subsection{RQ4: Ablation Studies}

\subsubsection{Training and Coordination}

Traffic-VL uses two complementary training stages. Offline human-informed RL learns
traffic-specific visual grounding and phase reasoning from semantically audited SynTra-V
images and expert labels, while online STPO-V refines the policy through multi-step
simulator interaction. We ablate offline GRPO, online STPO-V, and asynchronous coordination,
evaluating Real-1296 accuracy, Jinan1 closed-loop ATT, and corridor emergency waiting time.

\begin{table}[t]
  \centering
  \caption{Ablation of training, decision interface, and asynchronous coordination (mean$\pm$SD over five seeds).}
  \label{tab:rq4}
  \small
  \setlength{\tabcolsep}{3pt}
  \renewcommand{\arraystretch}{1.08}
  \begin{adjustbox}{max width=\linewidth}
  \begin{tabular}{@{}L{0.49\linewidth} rrr@{}}
    \toprule
    \textbf{Variant} & \textbf{Acc. (\%)} & \textbf{ATT (s)} & \textbf{AEWT (s)} \\
    \midrule
    \textbf{Full model (two-stage + coordination)} & $\mathbf{82.1\pm1.4}$ & $\mathbf{269.7\pm3.1}$ & $\mathbf{8.7\pm0.7}$ \\
    \midrule
    SFT-only (same backbone and data)  & $75.8\pm0.9$ & $288.9\pm4.1$ & -- \\
    SFT + STPO-V                       & $79.8\pm1.2$ & $278.6\pm3.6$ & -- \\
    Image-to-text bottleneck (same VLM/data) & $73.8\pm0.9$ & $291.1\pm4.0$ & -- \\
    Offline GRPO + stepwise PPO        & $80.2\pm1.2$ & $277.8\pm3.6$ & -- \\
    Offline-only (no STPO-V)            & $79.3\pm1.3$ & $281.3\pm3.8$ & -- \\
    Online-only (no offline GRPO)      & $72.2\pm1.5$ & $297.0\pm4.4$ & -- \\
    No asynchronous coordination       & $81.9\pm0.8$ & $269.9\pm3.4$ & $12.0\pm1.1$ \\
    \bottomrule
  \end{tabular}
  \end{adjustbox}
\end{table}

Table~\ref{tab:rq4} shows that both training stages are necessary. Under the same backbone,
data, and training budget, SFT-only reaches 75.8\% accuracy and 288.9\,s ATT; adding STPO-V
improves these values to 79.8\% and 278.6\,s, respectively. Offline GRPO provides a stronger
starting policy without token-level reasoning supervision.

We further isolate the decision interface with a matched image-to-text bottleneck using the
same Qwen3-VL backbone, SynTra-V images, and training data. This variant first verbalizes the
ordered views as lane-level traffic state and then selects a phase from that text and the
same candidate phase descriptions, without retaining the visual tokens in the decision
stage. It reaches 73.8\% accuracy and 291.1\,s ATT. The gap reflects the difficulty of
preserving the correspondence between visual lanes and phase descriptions through an
intermediate textual state.

Replacing STPO-V with stepwise PPO reaches 80.2\% accuracy and 277.8\,s ATT, further
supporting the trajectory-level objective. Removing offline GRPO
yields an online-only policy with 72.2\% accuracy and 297.0\,s ATT, indicating that
closed-loop exploration alone does not reliably acquire traffic-specific visual semantics
and phase knowledge. The offline-only variant improves to 79.3\% accuracy and 281.3\,s ATT,
but remains below the full model at 82.1\% and 269.7\,s, demonstrating the additional value
of long-horizon policy refinement.
Asynchronous coordination mainly benefits network-level control. Its removal changes
single-intersection results little, to 81.9\% accuracy and 269.9\,s ATT, but increases
corridor emergency waiting time from 8.7\,s to 12.0\,s. Directional traffic information
therefore helps upstream intersections prepare for approaching vehicles before they enter
the local view.

\subsubsection{Visual Domain Adaptation}

The training procedure deliberately separates visual adaptation from online policy
optimization. During offline learning, the vision encoder is updated on photorealistic
SynTra-V imagery to learn surveillance-style traffic appearance, including viewpoint,
illumination, weather, and occlusion variation. During online STPO-V, the simulator instead
provides deterministic Blender renders, and the vision encoder is frozen while the remaining
policy parameters are optimized. This prevents online optimization from overwriting visual
representations learned from photorealistic data with simulator-specific appearance.
We evaluate this design by unfreezing the vision encoder during online STPO-V and by
removing photorealistic stylization from SynTra-V. The variants are tested on stylized Jinan
closed-loop control, real surveillance frames, and OmniTraffic, whose observations are raw
simulator renders.

\begin{table}[t]
  \centering
  \caption{Effect of visual-domain adaptation on Jinan1, Real-1296, and OmniTraffic (mean$\pm$SD over five seeds).}
  \label{tab:rq5}
  \small
  \setlength{\tabcolsep}{3pt}
  \renewcommand{\arraystretch}{1.08}
  \begin{adjustbox}{max width=\linewidth}
  \begin{tabular}{@{}L{0.46\linewidth} rrr@{}}
    \toprule
    \textbf{Variant} & \multicolumn{1}{c}{\textbf{Jinan1}} &
    \multicolumn{1}{c}{\textbf{Real-1296}} &
    \multicolumn{1}{c}{\textbf{Omni}} \\
    & \textbf{ATT (s)} & \textbf{Acc. (\%)} & \textbf{Acc. (\%)} \\
    \midrule
    \textbf{Traffic-VL (full)} & $\mathbf{269.7\pm3.1}$ & $\mathbf{82.1\pm1.4}$ & $72.0\pm1.0$ \\
    \midrule
    Vision encoder unfrozen & $284.6\pm4.0$ & $73.4\pm1.5$ & $72.3\pm1.1$ \\
    No SynTra-V stylization & $295.6\pm4.6$ & $66.0\pm1.6$ & $\mathbf{73.1\pm0.7}$ \\
    \bottomrule
  \end{tabular}
  \end{adjustbox}
\end{table}

Table~\ref{tab:rq5} supports both design choices. Updating the vision encoder online raises
Jinan ATT from 269.7\,s to 284.6\,s and reduces real-frame accuracy from 82.1\% to 73.4\%.
This result suggests that exposure to raw simulator imagery during online learning weakens
the real-style visual representation acquired offline. Removing photorealistic stylization
has a larger effect, increasing ATT to 295.6\,s and lowering real-frame accuracy to 66.0\%.
In contrast, all variants achieve approximately 72\% accuracy on OmniTraffic. Because
OmniTraffic uses raw simulator renders, it does not expose the appearance gap that
photorealistic synthesis and encoder freezing are designed to address.

\begin{figure}[!b]
  \centering
  \includegraphics[width=0.93\linewidth]{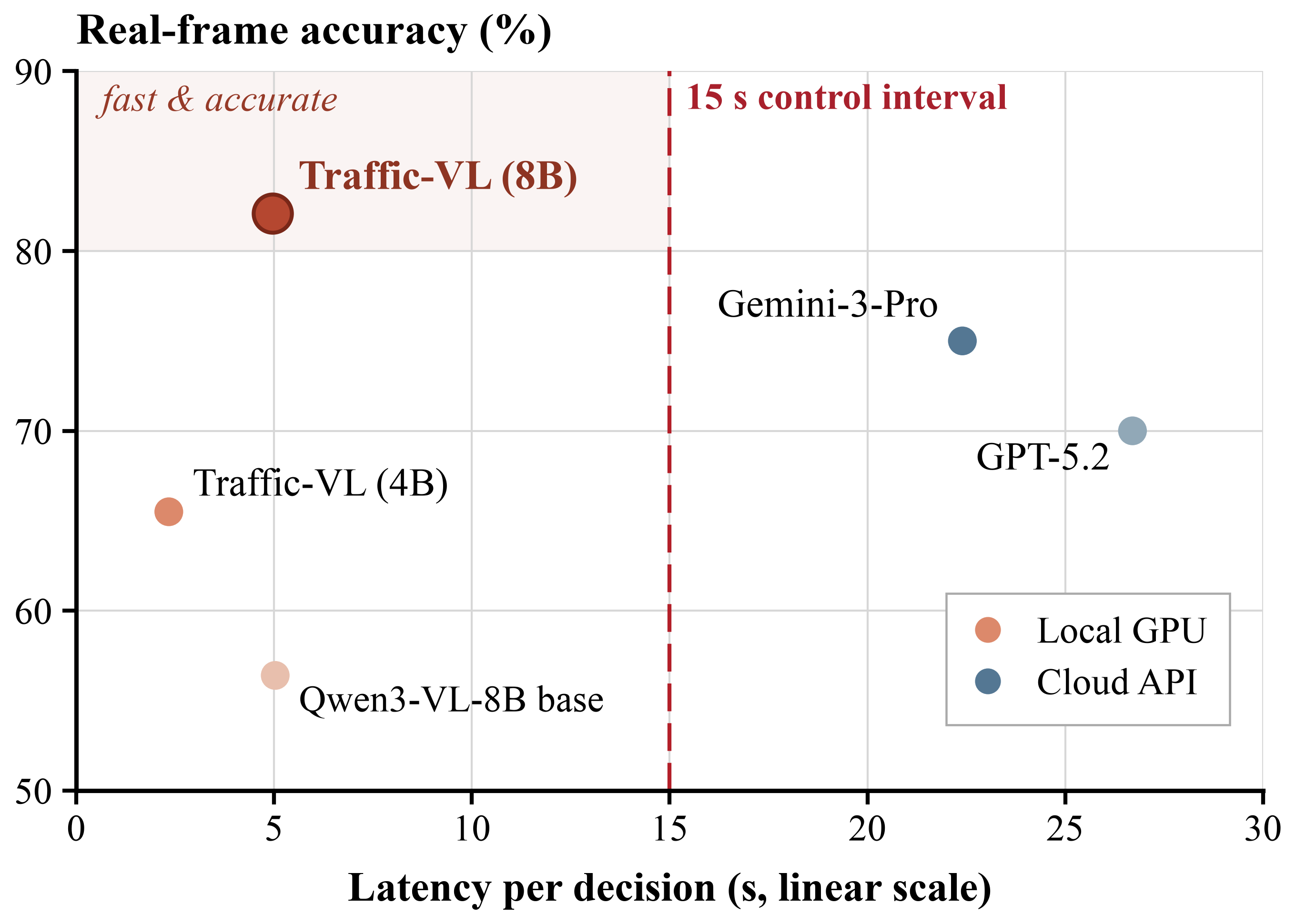}
    \caption{Real-frame accuracy versus decision latency for local models and cloud APIs, with the 15\,s control deadline marked.}
  \label{fig:qualitative}
\end{figure}

\subsection{RQ5: Inference Efficiency and the Latency-Accuracy Trade-off}
We evaluate whether Traffic-VL can satisfy the operational control deadline while
maintaining sufficient decision accuracy. The deployed system follows a nominal 15\,s
decision cycle, defining the maximum time from receiving the current multi-view
observation to issuing the next phase recommendation. Lower latency remains preferable
because traffic conditions may continue to evolve within this interval.

All local latency measurements are obtained on a single NVIDIA A800 GPU with 80\,GB
memory. We use unquantized BF16 inference with a batch size of one. Each decision takes
four $512\times512$ images, corresponding to approximately 1.3K visual tokens after
tokenization, and the generated response is capped at 1024 tokens. We perform 3 warm-up
runs followed by 20 timed runs. The reported end-to-end latency includes image
preprocessing, visual encoding, and autoregressive response generation, with GPU
synchronization applied before and after each measurement. Traffic-VL-8B requires 4.96\,s per decision, while Traffic-VL-4B
requires 2.34\,s. The unfine-tuned Qwen3-VL-8B base model requires 5.03\,s, indicating that
our reinforcement-learning adaptation introduces negligible inference overhead relative
to the same backbone. Both Traffic-VL variants therefore satisfy the 15\,s deadline in
this server-class local setting. 

For proprietary models, we report the end-to-end cloud API latency observed under our
evaluation environment. Gemini 3 Pro and GPT-5.2 require 22.4\,s and 26.7\,s per decision,
respectively. These values include network transmission, service queuing, and provider-side
inference. Under our tested environment, both cloud APIs exceed the 15\,s
deadline and achieve lower real-frame accuracy than Traffic-VL-8B.

Fig.~\ref{fig:qualitative} also illustrates the effect of model scale. Although
Traffic-VL-4B is faster, it achieves only 65.5\% real-frame accuracy, compared with 82.1\%
for Traffic-VL-8B, and follows the required output format less reliably after training.
Because both local variants meet the deadline on the A800, model capacity rather than
latency becomes the primary constraint in this setting. Traffic-VL-8B therefore provides
the better latency-accuracy trade-off for our server-based deployment.

\section{Real-World Deployment}

\subsection{From Advance Dispatch to Constrained Online Control}

\begin{figure}[b]
    \centering
    \includegraphics[width=\linewidth]{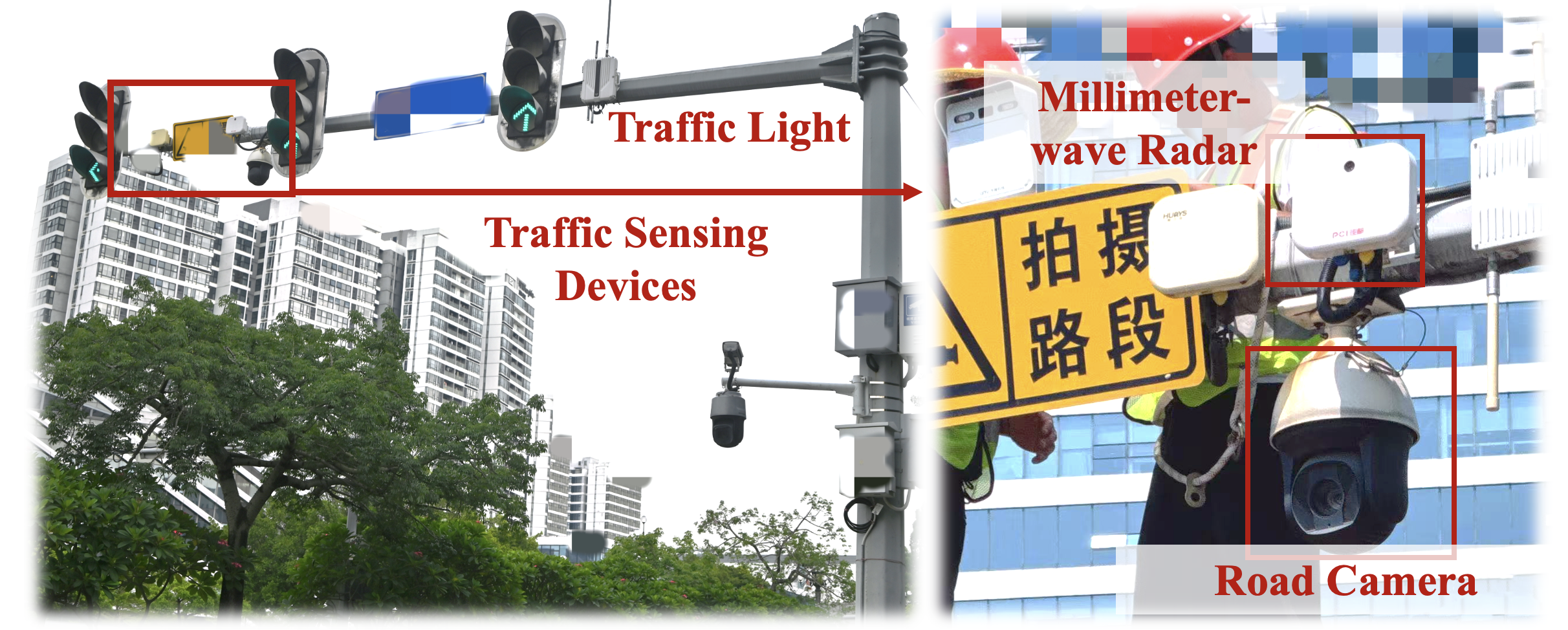}
    \caption{Field sensing infrastructure with roadside cameras, millimeter-wave radar, and the existing traffic signal controller.}
    \label{fig:field_infrastructure}
\end{figure}

Traffic-VL is deployed on the same traffic-management platform and ten interconnected
arterial intersections used in the Traffic-R1 field study. These intersections serve more
than 55,000 drivers daily. As shown in Fig.~\ref{fig:field_infrastructure}, the existing
platform integrates roadside cameras and millimeter-wave radar with the signal controller.

Traffic-R1 relied on numerical traffic states produced by these sensing devices and a
fixed, intersection-specific phase schema. Although lane aggregation, contextual prompts,
and additional fine-tuning adapted the model to the deployment sites, its decisions
remained tied to predefined state and action interfaces. Routine operation therefore used
\emph{advance dispatch}: traffic demand was forecast from historical observations,
Traffic-R1 generated a preliminary timing plan, and engineers reviewed and revised the plan
before implementation. This workflow supported offline planning but limited the model's
ability to respond directly to evolving traffic conditions.

Traffic-VL instead processes synchronized frames from the existing surveillance cameras
and interprets the approved signal plan through natural-language phase descriptions. It
does not require per-intersection lane-count encoding or a topology-specific output head,
allowing the model to recommend a preferred phase from current visual demand rather than
from an advance traffic forecast.

\begin{figure}[t]
    \centering
    \includegraphics[width=\linewidth]{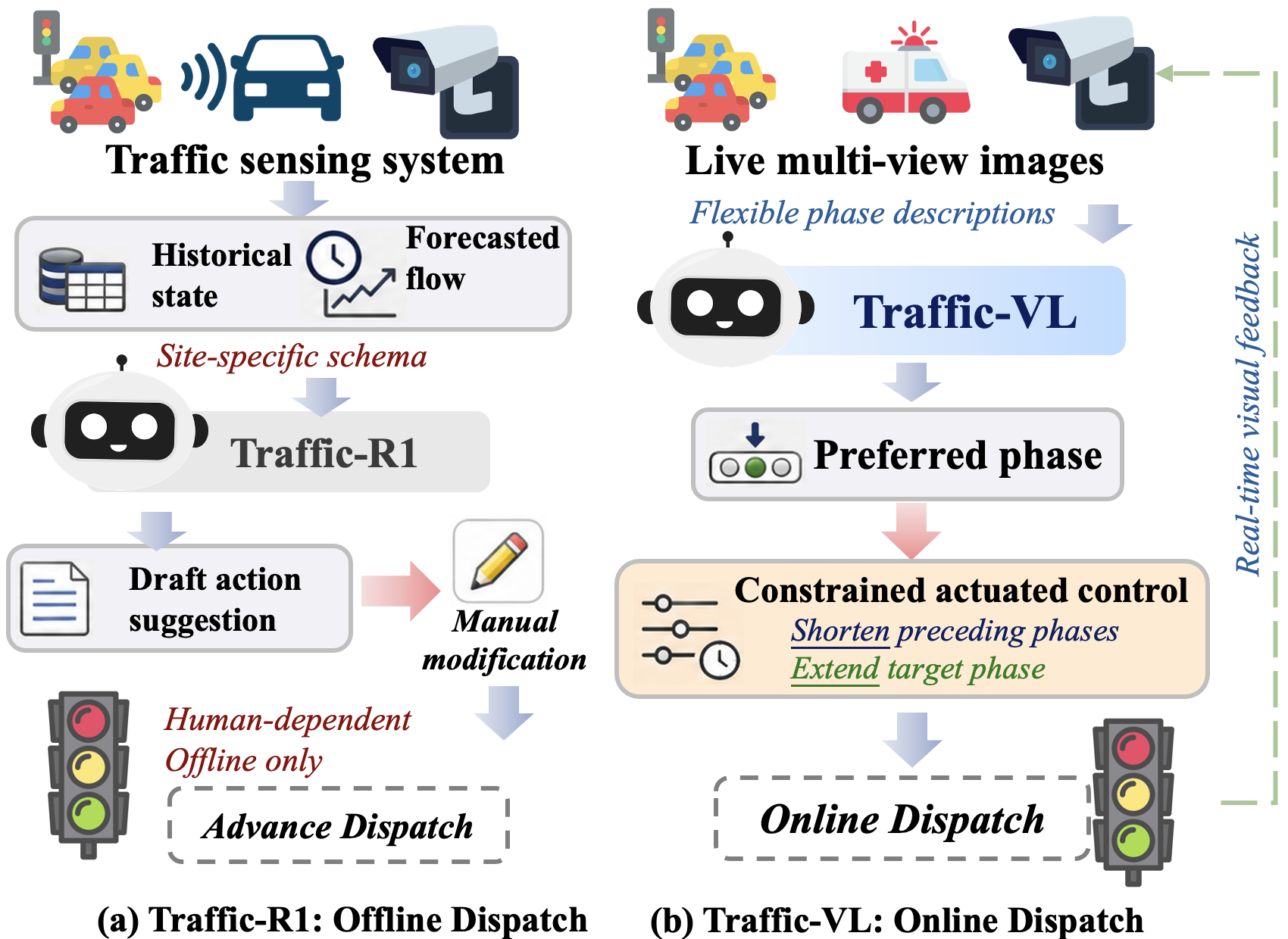}
    \caption{Deployment workflows: (a) Traffic-R1 advance dispatch with manual revision; (b) Traffic-VL constrained online phase-priority control.}
    \label{fig:deployment_workflow}
\end{figure}

Field operation follows a certified phase sequence with minimum and maximum green times,
pedestrian service, and yellow and all-red clearance intervals. Traffic-VL is therefore
connected to the existing signal controller through a phase-priority interface: the model
recommends which phase should receive priority, and the controller applies the admissible
timing adjustment within these constraints. If the recommended phase is currently green,
the controller extends it up to the maximum green time; otherwise, it terminates the
current phase once its minimum green has elapsed and advances along the approved sequence,
serving intermediate phases at minimum green until the recommended phase is reached.
Fig.~\ref{fig:deployment_workflow} summarizes the resulting change in deployment.
Traffic-R1 produces a manually revised advance plan from forecast structured states,
whereas Traffic-VL uses live visual observations to guide constrained online phase-priority
control with continuous feedback.

The deployed system runs Traffic-VL-8B with BF16 inference on the centralized
traffic server. Traffic-VL preserves the end-to-end mapping from raw observations to phase
recommendations, while the existing controller provides constraint-preserving execution.
Its visual grounding identifies where traffic demand is accumulating, and the multi-step
policy learned through STPO-V accounts for how current adjustments affect subsequent
traffic states. This combination replaces offline advance planning with bounded,
traffic-responsive online adaptation.

\subsection{Three-Week Field Evaluation}

We operated Traffic-VL for three consecutive weeks on the same ten-intersection network,
evaluating fifteen weekday evening peaks from 16:00 to 19:30. Manual operation and
Traffic-R1 are compared using the results of the Traffic-R1 field
study~\cite{zou2025trafficr1}, which used the same intersections, roadside measurement
system, and evaluation window.

A queued vehicle was defined as a vehicle traveling below $0.1\,\mathrm{m/s}$ on an
incoming approach. At each sampling time, the queue at an intersection was the total
number of queued vehicles across its incoming approaches. Average queue was computed over
all sampling times, intersections, and evaluation periods. Maximum queue was obtained by
first identifying the peak queue in each intersection-period and then averaging these
peaks. Operator workload measures the daily time spent preparing, reviewing, and adjusting
signal timing plans.

\begin{table}[t]
  \centering
  \caption{Three-week evaluation across ten intersections (queue lengths: mean$\pm$SD; timing-plan workload: h/day).}
  \label{tab:deploy}
  \small
  \setlength{\tabcolsep}{2.5pt}
  \renewcommand{\arraystretch}{1.08}
  \begin{adjustbox}{max width=\linewidth}
  \begin{tabular}{@{}L{0.29\linewidth} L{0.20\linewidth} ccc@{}}
    \toprule
    \textbf{Operating mode} & \textbf{Dispatch} &
    \textbf{Avg.\ Q$\downarrow$} &
    \textbf{Max.\ Q$\downarrow$} &
    \textbf{Plan h$\downarrow$} \\
    \midrule
    Manual control
      & Manual          & $34.5\pm3.6$ & $50.3\pm4.8$ & 2+ \\

    Traffic-R1
      & Advance         & $31.3\pm2.9$ & $48.1\pm3.9$ & 0.5+ \\

    \textbf{Traffic-VL}
      & \textbf{Online}
      & $\mathbf{30.9\pm3.1}$ & $\mathbf{47.5\pm4.2}$ & $\boldsymbol{\approx 0}$ \\
    \midrule
    \multicolumn{2}{@{}l}{\textit{Improvement vs.\ manual}}
      & $\downarrow$10.4\% & $\downarrow$5.6\% & $\downarrow\!\approx$100\% \\
    \bottomrule
  \end{tabular}
  \end{adjustbox}
\end{table}

As shown in Table~\ref{tab:deploy}, Traffic-VL reduces average and maximum queue lengths by
10.4\% and 5.6\%, respectively, relative to historical manual operation at the same
intersections. It also nearly eliminates
routine timing-plan workload by replacing advance plan preparation with online phase
adjustment. In comparison, Traffic-R1 achieved reductions of 9.3\% and 4.4\% while
still requiring approximately 0.5 operator hours per day for plan review.
Across the 15 evening-peak periods, the bootstrap 95\% confidence intervals for the
reductions are 5.1--15.2\% for average queue and 0.9--10.1\% for maximum queue. The median and 95th-percentile end-to-end field latencies are 6.3\,s and 9.2\,s, respectively.

The field results demonstrate that Traffic-VL can convert live visual observations into
useful phase priorities for constrained online control. Beyond improving traffic flow and
reducing routine workload, it removes dedicated radar or V2X measurements from the model
input and eliminates the per-intersection pipeline used to construct structured traffic
states. Existing roadside cameras can therefore support more scalable and responsive
deployment across heterogeneous intersections.

\section{Limitations and Future Work}

Traffic-VL currently conditions each decision on synchronized multi-view still images
rather than video. This protocol matches Traffic-R1 and established TSC baselines, allowing
us to isolate the effect of replacing structured states with visual observations. Although
STPO-V learns the long-term consequences of successive decisions, the model receives no
explicit motion history at each inference step. Short surveillance clips could reveal
whether traffic demand is accumulating or dissipating, improve reasoning under temporary
occlusion, and provide stronger evidence for abnormal events. However, video substantially
increases the visual-token sequence, memory consumption, and inference latency. Developing
an efficient temporal representation without compromising real-time deployment is
therefore an important direction for future work.

Traffic-VL also inherits residual perceptual errors and hallucinations from its
vision-language backbone. In rare safety-critical scenes, it may miss an emergency vehicle
or incorrectly assign priority to an ordinary vehicle. The existing signal controller
prevents infeasible signal transitions, but it cannot eliminate every suboptimal decision
caused by incorrect scene interpretation. This risk is greater in multi-intersection
coordination, where an erroneous flow-direction message may propagate to neighboring
controllers. Traffic-VL therefore remains integrated with certified signal-control
constraints rather than directly actuating unrestricted signal transitions. Future work will explore
stronger visual grounding, calibrated uncertainty, temporal-consistency checks, and
confidence-aware communication. With further validation, these mechanisms may support
progressively more autonomous deployment.

\section{Conclusion}

This work presented Traffic-VL, an 8B vision-language model that maps synchronized
multi-view surveillance images and natural-language phase descriptions directly to traffic
signal decisions. Traffic-VL replaces intersection-specific structured states with
end-to-end vision-language reasoning, supported by the SynTra-V semantics-preserving
Sim2Real data engine, two-stage offline and online reinforcement learning, a
natural-language phase interface, and asynchronous coordination. Operating on raw images,
it matches the structured-state Traffic-R1 on five zero-shot closed-loop benchmarks,
handles emergency events and camera occlusion more reliably than perception-to-structure
pipelines, and transfers to unseen topologies and real surveillance frames within the
real-time inference budget. In a three-week deployment across ten interconnected intersections,
Traffic-VL reduced average queue length by 10.4\% relative to historical manual operation
at the same sites, showing that camera-based reasoning can support effective constrained
online control in a physical traffic network.

More broadly, our results suggest that real-world traffic signal control can be formulated
as a unified vision-language decision problem rather than a sequence of independently
engineered perception, state-construction, and control modules. Preserving raw visual
evidence until the decision stage allows multimodal foundation models to jointly ground
physical traffic demand in symbolic control semantics, reducing dependence on dedicated
sensing hardware and per-intersection dynamic-state engineering. Extending this paradigm with efficient
temporal perception, calibrated uncertainty, and progressively validated autonomy offers a
promising path toward generalizable and scalable foundation models for real-world urban
infrastructure control.

\FloatBarrier
\bibliographystyle{IEEEtran}
\bibliography{refs}

\clearpage
\appendices
\section{Information-Preservation Rationale}
\label{app:theory}

We formalize the information-preservation principle underlying the design rationale in
Section~\ref{sec:design-rationale}. Let $S$ denote the latent traffic state, $O$ the raw
multi-view observation, and $Z=T(O)$ the structured state produced by a fixed deterministic
extractor $T$. At a decision step, the controller selects a phase $A\in\mathcal P$ and
receives utility $u(S,A)$. The candidate phases and operational context are treated as part
of the decision context and held fixed in the comparison.

For an observation variable $X$, let $\Pi_X$ be the set of possibly randomized decision
rules measurable with respect to $X$. Its optimal decision value is
\begin{equation}
V^*(X)
=
\sup_{\pi\in\Pi_X}
\mathbb E\!\left[u\big(S,\pi(X)\big)\right].
\label{eq:bayes_value}
\end{equation}
This definition compares observation interfaces through the policy classes they support.

\smallskip\noindent\textbf{Proposition 1 (Information ceiling).}
For every deterministic extractor $T$,
\begin{equation}
V^*(O)\ge V^*(T(O)).
\label{eq:information_ceiling}
\end{equation}

\smallskip\noindent\emph{Proof.}
Take any rule $\pi_Z\in\Pi_Z$. Its lifted rule
$\pi_O(o)=\pi_Z(T(o))$ belongs to $\Pi_O$ and induces the same action distribution and
expected utility. Composition with $T$ therefore embeds every $Z$-based rule into the
$O$-based policy class. Taking the supremum over the two classes yields
Eq.~\ref{eq:information_ceiling}. \hfill$\square$

Two cases follow from Eq.~\ref{eq:information_ceiling}. Equality holds when an optimal
raw-observation policy factors through $T$, so that $T(O)$ is decision-sufficient for phase
selection. The inequality is strict when no optimal raw-observation policy admits such a
factorization. A direct instance is \emph{schema aliasing}: two positive-probability visual
situations map to the same structured state but have different optimal phases. Every
$Z$-based policy must use the same action distribution in both situations and therefore
cannot attain both raw-observation optima. Once this distinction is removed by $T$, increasing
the capacity of the downstream controller cannot recover it.

For Traffic-VL, the implication is direct. Retaining the multi-view observation until phase
selection preserves the full feasible decision class, whereas a fixed extraction of counts,
occupancies, or detector outputs restricts that class to decisions expressible through the
schema. A structured interface reaches the same ceiling precisely when it preserves the
evidence required by the optimal decision. Model capacity and training determine the value
realized within each policy class; the information ordering established above remains fixed.
The same lifting argument applies when history or additional decision context is included in
$O$.

\section{SynTra-V Generation and Audit Protocol}
\label{app:audit}

This section traces the complete path from a simulator-rendered view to an accepted
photorealistic observation. We first specify the generation inputs and prompt, then describe
the four independent semantic audits and their joint acceptance rule, and finally report
calibration against blind human judgments.

\subsection{Generation Inputs}

For each synchronized camera view, SynTra-V uses the multi-image editing interface of
\texttt{gpt-image-2} to generate a $2048\times2048$ image. The generator receives three
pixel-aligned inputs in a fixed order: the native Blender rendering, a vehicle-control image,
and a lane-control image. The depth reference is reserved for post-generation auditing and
does not condition generation. Accepted images are resized to $512\times512$ before
being passed to the controller.

\subsection{Generation Prompt Template}

Generation is performed in a single pass, with vehicle count, city style, and weather or
lighting supplied through scene-specific fields. Figure~\ref{fig:generation-prompt} records
the attachment order and the prompt.

\begin{figure*}[!t]
\begin{promptbox}{Generation Input Contract}
\footnotesize
\textbf{Image 1 -- source RGB:} the native Blender rendering to be transformed.
\textbf{Image 2 -- $C_{\mathrm{veh}}$:} the source view overlaid with one near-to-far
numbered red box per vehicle and a green arrow obtained by projecting that vehicle's 3D
heading; annotations scale with object size.
\textbf{Image 3 -- $C_{\mathrm{lane}}$:} the complete lane lines, stop lines, crosswalks,
and turn arrows, with intensity decreasing mildly with distance.
\textbf{Dynamic fields:} \texttt{\{N\}}, \texttt{\{CITY\_STYLE\}}, and
\texttt{\{WEATHER\_LIGHTING\}}.
\end{promptbox}

\begin{promptbox}{Final SynTra-V Generation Prompt}
\footnotesize
Turn Image 1, a 3D-rendered road intersection, into a photorealistic real urban photograph,
never a 3D render, CGI image, or game scene. Apply \texttt{\{CITY\_STYLE\}} and
\texttt{\{WEATHER\_LIGHTING\}}. Add a realistic streetscape beyond the road, including
sidewalks, trees, lamp posts, shops, and a few pedestrians. Do not add a timestamp, watermark,
text overlay, or visible control annotation.

Image 2 marks each of the exactly \texttt{\{N\}} vehicles with a numbered red box and a green
arrow showing where its front points. Image 3 maps the painted road markings, brighter when
near and dimmer when far. All three images are pixel-aligned views from the same camera. Never
draw the boxes, indices, arrows, or lane-control map in the output.

\textbf{Vehicles.} Produce exactly \texttt{\{N\}} vehicles, one per red box, at the same
position and scale. Preserve every vehicle's lane occupancy, spacing, footprint, and facing
direction. Orientation is invariant: if Image 1 shows the front, the output shows the front;
if it shows the rear, the output shows the rear. Never exchange front and rear, turn a vehicle
around, create a wrong-way vehicle, move a vehicle across a lane, add a vehicle, or remove one.
Retexture each existing vehicle in place using photorealistic paint, glass, and wheels. Replace
all pastel placeholder colors with a natural traffic palette, dominated by white, black,
silver, and gray with a few muted colors and an occasional taxi. Vehicles should look driven
rather than showroom-new, with subtle road dust on lower bodywork, light brake dust on wheels,
and faint glass smudges. Apply mild directional motion blur only to vehicles already moving;
queued vehicles remain sharp.

\textbf{Road markings and surface.} Reproduce every lane line, lane divider, centerline, stop
line, crosswalk, and turn arrow at the location specified by Image 3. Preserve lane count and
connectivity; do not add, remove, merge, or laterally shift markings. Keep dashed markings
dashed as shown in Image 1, even when the control channel uses continuous traces for clarity.
Lane markings remain visible to the far end of the road, becoming thinner with perspective
without fading away. Markings occluded by vehicles remain occluded, and no arrow or marking is
added beyond the stop line unless Image 3 contains it. Render naturally worn paint with faded
edges and small scuffs. Rain may make the paint glossy but must not erase it. Render aged,
non-uniform asphalt with subtle tonal patches, repairs, sealant cracks, tire-wear tracks, and
faint oil stains where vehicles queue.

\textbf{Photographic appearance and remaining roadway.} The frame must read as a genuine
surveillance-style photograph with natural optics, mild sensor grain, and physically
consistent shadows, never as a clean computer render. Keep all road surface outside the
\texttt{\{N\}} boxes free of vehicles, including the far background; do not invent distant or
horizon traffic. Final check: exactly \texttt{\{N\}} vehicles, each inside its box and facing
along its arrow with front and rear unchanged; lane markings match Image 3; the far road
contains no added traffic.
\end{promptbox}
\caption{SynTra-V generation inputs and prompt template, with aligned vehicle/lane constraints and scene-specific city, weather, and lighting fields.}
\label{fig:generation-prompt}
\end{figure*}

\subsection{Independent Audit Prompts}

Each generated candidate is then evaluated by four independent Gemini 3 Pro calls, aligned
with the Box, Geometric, Depth, and Movement references introduced in
Section~\ref{sec:syntra}. Every call receives the relevant pixel-aligned Blender reference,
the generated RGB image, and an auxiliary result extracted from that image. YOLO11n,
morphological extraction, monocular depth estimation, and crop-level heading recognition
construct these auxiliary results; Gemini combines them with the images to issue the audit
verdict.

Figure~\ref{fig:audit-prompts} gives the shared system instruction and the four task-specific
prompts. The image indices in each prompt define the attachment order for that call.

\begin{figure*}[!t]
\centering
\begin{promptbox}{Shared System Prompt}
You are a strict semantic auditor for a traffic-scene image-generation pipeline. Treat the
Blender-derived reference as ground truth and evaluate control-relevant consistency rather
than pixel-level similarity. Permit changes in city style, weather, illumination, texture,
and rendering appearance. Use only the supplied images and auxiliary evidence. If the inputs
do not support a reliable decision, return \texttt{UNCERTAIN}. Return exactly one JSON object:
\texttt{\{"verdict": "PASS|FAIL|UNCERTAIN", "reason": "<one concise sentence>",
"mismatches": ["<item>"]\}}.
\end{promptbox}

\noindent\begin{minipage}[t]{0.49\textwidth}
\vspace{0pt}
\begin{promptbox}{Vehicle-Layout Audit}
\textbf{Input order:} Image 1 -- Blender Box reference; Image 2 -- generated RGB image;
Image 3 -- YOLO11n boxes detected from Image 2.\par
\textbf{User prompt:} Audit vehicle-layout preservation. Match the reference boxes in Image 1
to the detected boxes in Image 3, using Image 2 to resolve missed or imprecise detections.
Compare vehicle count, approximate box centers and sizes, lane assignment, and relative
front-to-back order. Return \texttt{FAIL} for a missing or duplicated vehicle, a vehicle moved
to another lane, or a major layout change. Do not fail solely because of small box offsets,
minor scale changes, or detector jitter.
\end{promptbox}

\begin{promptbox}{Geometric-Structure Audit}
\textbf{Input order:} Image 1 -- Blender Geometric reference used to construct
$C_{\mathrm{lane}}$; Image 2 -- generated RGB image; Image 3 -- morphological lane-line and
turn-arrow extraction from Image 2.\par
\textbf{User prompt:} Audit preservation of road and lane geometry. Compare the number and
connectivity of approaches, lane count, lane boundaries, centerlines, and turn arrows across
the three inputs. Use Image 2 to distinguish true structural changes from markings hidden by
vehicles or weakly recovered in Image 3. Return \texttt{FAIL} for changed topology, removed or
added lanes, incompatible turn arrows, or visibly displaced road boundaries. Do not penalize
texture changes or plausible occlusion.
\end{promptbox}
\end{minipage}\hfill
\begin{minipage}[t]{0.49\textwidth}
\vspace{0pt}

\begin{promptbox}{Depth-Ordering Audit}
\textbf{Input order:} Image 1 -- pixel-aligned Blender Depth reference; Image 2 -- generated
RGB image; Image 3 -- monocular depth estimate computed from Image 2.\par
\textbf{User prompt:} Audit ordinal depth preservation; absolute depth scale need not match.
Compare near-to-far ordering for vehicles and major road regions, foreground-background
occlusion, and perspective-consistent overlap. Treat a monotonic rescaling of Image 3 as
valid. Return \texttt{FAIL} for reversed object order, impossible occlusion, or a major depth
layout inconsistent with Image 1. Return \texttt{UNCERTAIN} when the RGB view lacks enough
visible cues for comparison.
\end{promptbox}

\begin{promptbox}{Vehicle-Heading Audit}
\textbf{Input order:} Image 1 -- Blender Movement reference; Image 2 -- generated RGB image;
Image 3 -- vehicle crops from Image 2 annotated with crop-level heading predictions.\par
\textbf{User prompt:} Audit vehicle heading and movement preservation. Associate vehicles
across the inputs by box position and lane. For each visible vehicle, compare its predicted
heading with the direction encoded in Image 1 and verify that it is compatible with the lane's
legal movement. Return \texttt{FAIL} for a reversed vehicle, a cross-lane orientation, or a
heading incompatible with the reference movement. Allow small yaw differences introduced by
photorealistic rendering.
\end{promptbox}
\end{minipage}
\caption{SynTra-V auditing prompts share a system instruction and independently check vehicle layout, geometry, depth order, and vehicle heading.}
\label{fig:audit-prompts}
\end{figure*}

\subsection{Acceptance Rule}

A candidate is accepted only when all four calls return \texttt{PASS}. A \texttt{FAIL} or
\texttt{UNCERTAIN} verdict rejects the candidate, after which SynTra-V resamples an image
from the same simulator state. The four JSON outputs are retained separately.

\subsection{Blind Human Calibration}

We calibrate the joint decision rule on 500 candidate views sampled before automatic
filtering. Three traffic-domain experts independently assess preservation along the same four
dimensions, and majority vote defines the reference label. Table~\ref{tab:audit_calibration}
shows 96.1\% macro agreement with the human labels, a 2.8\% false-accept rate, and per-audit F1
above 0.94. Inter-expert agreement reaches Fleiss' $\kappa=0.93$. Constructing the final
5,000-sample corpus requires 6,420 candidates, corresponding to a 77.9\% acceptance rate and
1.28 candidates per accepted sample.

\begingroup

\begin{table}[t]
  \centering
  \caption{Blind human-expert calibration of the SynTra-V audit on 500 candidate views before filtering.}
  \label{tab:audit_calibration}
  \small
  \setlength{\tabcolsep}{4pt}
  \begin{adjustbox}{max width=\linewidth}
  \begin{tabular}{@{}L{0.68\linewidth} r@{}}
    \toprule
    \textbf{Audit statistic} & \textbf{Value} \\
    \midrule
    Candidate views                            & 500 \\
    Human traffic experts                      & 3 \\
    Macro agreement across four audits         & 96.1\% \\
    False accept / false reject                & 2.8\% / 4.4\% \\
    Per-audit F1 range                         & 0.942--0.965 \\
    Expert agreement (Fleiss' $\kappa$)        & 0.93 \\
    Full-corpus acceptance rate                & 77.9\% (5,000/6,420) \\
    \bottomrule
  \end{tabular}
  \end{adjustbox}
\end{table}

\endgroup

\section{Asynchronous Coordination Protocol}
\label{app:async_algorithm}

This section specifies the asynchronous coordination mechanism introduced in
Section~\ref{sec:coordination}. The presentation proceeds from the local state, action, and
message interfaces to the symmetric half-step schedule and message lifecycle. The protocol
follows Traffic-R1, with multi-view images and topology-conditioned phase descriptions in
place of structured textual observations.

\subsection{Local Inputs, Actions, and Messages}

Let the road network be a directed graph
\(G=(\mathcal{I},\mathcal{E})\), where \(\mathcal{I}\) is the set of signalized
intersections and \(\mathcal{E}\) represents their traffic-flow connectivity. The
intersections are partitioned by positional parity into two disjoint groups,
\(\mathcal{I}_1\) and \(\mathcal{I}_2\), such that
\(\mathcal{I}_1\cup\mathcal{I}_2=\mathcal{I}\). Adjacent intersections are assigned to
different groups whenever permitted by the network topology.

At intersection \(i\), the local policy input is
\begin{equation}
x_{i,t}
=
\left(
o_{i,t},
\mathcal{D}_i,
\mathcal{M}_{i,t}
\right),
\end{equation}
where \(o_{i,t}\) is the synchronized multi-view observation,
\(\mathcal{D}_i=\{D_i(p)\mid p\in\mathcal{P}_i\}\) contains the candidate phase
descriptions, and \(\mathcal{M}_{i,t}\) is the set of messages received from neighboring
intersections during the preceding inner step. Traffic-VL generates a structured response
\begin{equation}
y_{i,t}\sim\pi_\theta(\cdot\mid x_{i,t}),
\qquad
a_{i,t}=\operatorname{Parse}(y_{i,t}),
\end{equation}
where \(a_{i,t}\) is valid only if it contains exactly one phase identifier in the
candidate set \(\mathcal{P}_i\). Because every phase in \(\mathcal{P}_i\) is
conflict-free by construction, phase legality is enforced through membership in this set.
If the response is invalid or unparsable, the controller retains its previous valid action:
\begin{equation}
\tilde a_{i,t}
=
\begin{cases}
a_{i,t}, & a_{i,t}\in\mathcal{P}_i,\\
\tilde a_{i,t-1}, & \text{otherwise}.
\end{cases}
\label{eq:invalid_action_fallback}
\end{equation}
Yellow and all-red clearance intervals are inserted by the simulator and are not selected
by the policy.

Each active intersection also produces a directional-flow message for its downstream
neighbors. We represent the message sent from intersection \(i\) toward neighbor \(j\) as
\begin{equation}
m_{i\rightarrow j,t}
=
\left(
i,j,t,\hat f_{i\rightarrow j,t}
\right),
\label{eq:coord_message}
\end{equation}
where \(i\) and \(j\) identify the source and receiver, \(t\) is the control-step
timestamp, and \(\hat f_{i\rightarrow j,t}\) summarizes the traffic flow expected to leave
intersection \(i\) toward \(j\). In implementation, this tuple is serialized into a
compact natural-language message and appended to the receiver's next policy input. The
message supplements the receiver's local visual observation rather than replacing it.

\subsection{Symmetric Half-Step Coordination}

Each control step is divided into two inner steps. Intersections in
\(\mathcal{I}_1\) act first, using messages produced by \(\mathcal{I}_2\) during the
preceding inner step. Their new messages are then delivered to intersections in
\(\mathcal{I}_2\), which act in the second inner step. Intersections outside the active
group retain their previous actions. Algorithm~\ref{alg:async_coordination} presents the
complete procedure.

\subsection{Message Lifecycle and Execution Scope}

The message buffer is replaced after each group activation rather than accumulated over the
rollout. Messages are consumed at the receiver's next activation and are therefore delayed
by at most one inner step. The evaluation assumes reliable, in-order delivery between
eligible neighboring intersections.

Only the active group invokes Traffic-VL; inactive intersections retain their latest valid
phases. Because directional-flow estimates reach the other group before it acts, downstream
controllers can anticipate traffic not yet visible in their local views. Inference remains
decentralized: each decision uses local images, local phase descriptions, and messages from
nearby intersections, without access to a global structured traffic state.

\begin{algorithm*}[t]
\caption{Asynchronous Traffic-VL Control with Symmetric Half-Step Communication}
\label{alg:async_coordination}
\small
\begin{algorithmic}[1]
\Require Road network \(G=(\mathcal{I},\mathcal{E})\);
         intersection groups \((\mathcal{I}_1,\mathcal{I}_2)\);
         simulator \(\mathrm{Env}\);
         policy \(\pi_\theta\);
         communication radius \(d_{\max}=2\,\mathrm{km}\);
         total control steps \(H\)
\Ensure Sequence of executed phase actions
        \(\{\tilde a_{i,t}\}_{i\in\mathcal I,t=1}^{H}\)

\State Initialize a valid action \(\tilde a_{i,0}\in\mathcal P_i\) for every
       \(i\in\mathcal I\)
\State Initialize the incoming message buffer
       \(\mathcal M^{2\rightarrow1}\gets\varnothing\)

\For{\(t=1,\ldots,H\)}
    \State \Comment{Inner step 1: activate \(\mathcal I_1\)}
    \State \((\mathcal A_1,\mathcal M^{1\rightarrow2})
       \gets
       \Call{ActivateGroup}
       {\mathcal I_1,\mathcal I_2,\mathcal M^{2\rightarrow1},t}\)
    \State \(\mathrm{Env}.\Call{Update}
       {\mathcal A_1,\{\tilde a_{j,t-1}\}_{j\in\mathcal I_2}}\)
       \Comment{\(\mathcal I_2\) retains previous actions}

    \State \Comment{Inner step 2: activate \(\mathcal I_2\)}
    \State \((\mathcal A_2,\mathcal M^{2\rightarrow1})
       \gets
       \Call{ActivateGroup}
       {\mathcal I_2,\mathcal I_1,\mathcal M^{1\rightarrow2},t}\)
    \State \(\mathrm{Env}.\Call{Update}
       {\mathcal A_2,\{\tilde a_{i,t}\}_{i\in\mathcal I_1}}\)
       \Comment{\(\mathcal I_1\) retains its latest actions}
\EndFor

\Procedure{ActivateGroup}
          {$\mathcal I_{\mathrm{act}},
             \mathcal I_{\mathrm{recv}},
             \mathcal M_{\mathrm{in}},t$}
    \State \(\mathcal A\gets\varnothing\)
    \State \(\mathcal M_{\mathrm{next}}\gets\varnothing\)

    \ForAll{\(i\in\mathcal I_{\mathrm{act}}\)}
        \State \(o_{i,t}\gets\mathrm{Env}.\Call{GetVisualObservation}{i}\)
        \State \(\mathcal D_i\gets
            \{D_i(p)\mid p\in\mathcal P_i\}\)
        \State \(\mathcal M_{i,t}\gets
            \mathcal M_{\mathrm{in}}[i]\)
        \State \(y_{i,t}\sim
            \pi_\theta(\cdot\mid o_{i,t},\mathcal D_i,\mathcal M_{i,t})\)
        \State \(a_{i,t}\gets\Call{Parse}{y_{i,t}}\)

        \If{\(a_{i,t}\in\mathcal P_i\)}
            \State \(\tilde a_{i,t}\gets a_{i,t}\)
        \Else
            \State \(\tilde a_{i,t}\gets\tilde a_{i,t-1}\)
            \Comment{Retain the previous valid action}
        \EndIf

        \State \(\mathcal A[i]\gets\tilde a_{i,t}\)

        \ForAll{\(j\in\mathcal N(i)\cap\mathcal I_{\mathrm{recv}}\)}
            \If{\(\operatorname{dist}(i,j)\le d_{\max}\)}
                \State \(\hat f_{i\rightarrow j,t}
                    \gets
                    \Call{EstimateDirectionalFlow}
                    {o_{i,t},j}\)
                \State \(m_{i\rightarrow j,t}
                    \gets
                    (i,j,t,\hat f_{i\rightarrow j,t})\)
                \State \(\mathcal M_{\mathrm{next}}[j]
                    \gets
                    \mathcal M_{\mathrm{next}}[j]
                    \cup\{m_{i\rightarrow j,t}\}\)
            \EndIf
        \EndFor
    \EndFor

    \State \Return \((\mathcal A,\mathcal M_{\mathrm{next}})\)
\EndProcedure
\end{algorithmic}
\end{algorithm*}

\FloatBarrier

\section{Statistical Reporting}
\label{app:statistics}

Table~\ref{tab:paired_uncertainty} reports mean$\pm$SD over the five paired demand seeds used
in Table~\ref{tab:rq1}. Paired two one-sided tests (TOST) with a pre-specified $\pm2\%$
relative margin support ATT/AWT equivalence between Traffic-VL and Traffic-R1 on every
network after Holm correction (all adjusted $p<0.05$).

\begingroup

\begin{table}[H]
  \centering
  \caption{Paired ATT/AWT over five demand seeds (mean$\pm$SD; seconds).}
  \label{tab:paired_uncertainty}
  \scriptsize
  \setlength{\tabcolsep}{3pt}
  \begin{adjustbox}{max width=\linewidth}
  \begin{tabular}{@{}l rr rr@{}}
    \toprule
    \multirow{2}{*}{\textbf{Network}} &
    \multicolumn{2}{c}{\textbf{Traffic-R1}} &
    \multicolumn{2}{c}{\textbf{Traffic-VL}} \\
    \cmidrule(lr){2-3}\cmidrule(l){4-5}
    & \textbf{ATT} & \textbf{AWT} & \textbf{ATT} & \textbf{AWT} \\
    \midrule
    Jinan1    & $270.34\pm3.18$ & $27.95\pm0.74$ & $269.71\pm3.26$ & $27.82\pm0.77$ \\
    Jinan2    & $239.53\pm2.71$ & $21.11\pm0.62$ & $239.81\pm2.79$ & $21.05\pm0.65$ \\
    Jinan3    & $238.03\pm2.56$ & $23.17\pm0.68$ & $237.95\pm2.63$ & $23.31\pm0.70$ \\
    Hangzhou1 & $277.83\pm3.42$ & $15.51\pm0.51$ & $277.56\pm3.50$ & $15.62\pm0.54$ \\
    Hangzhou2 & $324.11\pm4.06$ & $33.14\pm0.91$ & $323.89\pm4.15$ & $33.12\pm0.94$ \\
    \bottomrule
  \end{tabular}
  \end{adjustbox}
\end{table}

\endgroup

\section{Baseline Definitions}
\label{app:baselines}

Table~\ref{tab:baselines} provides a unified inventory of the baselines used in
Section~\ref{sec:experiments}, including their method category, observation interface, and
role in the comparison. The VLMLight (original) entry denotes the structured-state
reference configuration specified in the experimental setup, whose reasoning and control
components receive simulator-provided traffic descriptions. VLMLight (adapted) retains
the scene-understanding and reasoning pipeline while receiving SynTra-V images.
Advanced-CoLight$\dagger$ uses a fine-tuned Grounding DINO detector and topology-specific
mappings to construct structured states for the original controller.

\begin{table*}[t]
  \centering
  \caption{Baseline taxonomy, observation interfaces, and roles in the comparison.}
  \label{tab:baselines}
  \small
  \setlength{\tabcolsep}{5pt}
  \begin{tabular}{@{}l l l L{0.46\textwidth}@{}}
    \toprule
    \textbf{Baseline} & \textbf{Category} & \textbf{Observation} & \textbf{Description} \\
    \midrule
    FixedTime~\cite{koonce2008traffic} & Rule-based & Structured-Oracle & Fixed-cycle plan with preset phase durations; no learning. \\
    MaxPressure~\cite{varaiya2013maxpressure} & Rule-based & Structured-Oracle & Greedily activates the phase of maximum pressure each step. \\
    MPLight~\cite{chen2020mplight} & RL & Structured-Oracle & Pressure-based deep RL with FRAP backbone and parameter sharing. \\
    CoLight~\cite{wei2019colight} & RL & Structured-Oracle & Graph-attention RL that coordinates with neighboring intersections. \\
    Advanced-CoLight~\cite{zhang2022advanced} & RL & Structured-Oracle & Strongest RL baseline; advanced traffic-state features over CoLight. \\
    AttendLight~\cite{oroojlooy2020attendlight} & RL & Structured-Oracle & Attention policy across intersection structures; evaluated zero-shot without target-intersection fine-tuning in the main comparison. \\
    LLMLight~\cite{lai2025llmlight} & LLM & Structured-Oracle (text) & LLM controller fine-tuned or prompted on textual per-lane state. \\
    CoLLMLight~\cite{yuan2026collmlight} & LLM & Structured-Oracle (text) & Cooperative LLM controller for multi-intersection coordination. \\
    Traffic-R1~\cite{zou2025trafficr1} & LLM & Structured-Oracle (text) & Our conference model; two-stage RL over structured traffic text. \\
    VLMLight (original)~\cite{wang2025vlmlight} & VLM & Structured-Oracle (text) & Structured-state reference configuration in this comparison; its reasoning and control components receive simulator-provided traffic descriptions. \\
    Qwen3-VL 8B (base)~\cite{bai2025qwen3vl} & VLM & Raw-Image & Unfine-tuned deployment backbone, prompted zero-shot. \\
    Gemini 3 Pro~\cite{deepmind2025gemini3pro} & VLM & Raw-Image & Frontier general-purpose VLM, evaluated zero-shot. \\
    \midrule
    VLMLight (adapted)~\cite{wang2025vlmlight} & Adapted & Raw-Image & Original scene-understanding and reasoning pipeline evaluated on the same SynTra-V observations as Traffic-VL. \\
    Advanced-CoLight$\dagger$~\cite{zhang2022advanced} & Adapted & Raw-Image$\to$Struct & A fine-tuned Grounding DINO detector supplies vehicle detections, which topology-specific mappings convert into queue and pressure states for the controller. \\
    \bottomrule
  \end{tabular}
\end{table*}

\FloatBarrier

\section{Full-Shot Comparison}
\label{app:fullshot}

The main evaluation emphasizes zero-shot transfer. For context, Table~\ref{tab:fullshot}
also reports conventional RL controllers retrained on each target benchmark under
Traffic-R1's full-shot protocol. Traffic-R1 and Traffic-VL remain zero-shot, with their
values reproduced from Table~\ref{tab:rq1}; the comparison therefore separates
benchmark-specific optimization from transfer performance.

\makeatletter
\setlength{\@dblfptop}{0pt}
\makeatother
\begin{table*}[!t]
  \centering
  \caption{Full-shot baseline comparison on five public benchmarks, with zero-shot Traffic-R1 and Traffic-VL as references (ATT/AWT in seconds).}
  \label{tab:fullshot}
  \small
  \setlength{\tabcolsep}{4pt}
  \begin{tabular}{@{}l l rr rr rr rr rr@{}}
    \toprule
    \multirow{2}{*}{\textbf{Method}} & \multirow{2}{*}{\textbf{Setting}} &
    \multicolumn{2}{c}{\textbf{Jinan1}} & \multicolumn{2}{c}{\textbf{Jinan2}} &
    \multicolumn{2}{c}{\textbf{Jinan3}} & \multicolumn{2}{c}{\textbf{Hangzhou1}} &
    \multicolumn{2}{c}{\textbf{Hangzhou2}} \\
    \cmidrule(l){3-4} \cmidrule(l){5-6} \cmidrule(l){7-8} \cmidrule(l){9-10} \cmidrule(l){11-12}
    & & ATT & AWT & ATT & AWT & ATT & AWT & ATT & AWT & ATT & AWT \\
    \midrule
    FixedTime        & Rule-based & 453.41 & 51.32 & 370.34 & 35.15 & 384.53 & 36.95 & 497.54 & 36.41 & 408.53 & 53.94 \\
    MaxPressure      & Rule-based & 274.34 & 32.04 & 246.35 & 22.56 & 245.66 & 24.31 & 289.55 & 21.52 & 349.53 & 67.52 \\
    MPLight          & Full-shot  & 310.54 & 50.45 & 270.14 & 48.03 & 272.37 & 42.91 & 319.32 & 44.05 & 365.42 & 69.85 \\
    CoLight          & Full-shot  & 272.44 & 41.09 & 250.41 & 38.59 & 248.84 & 39.94 & 294.61 & 42.18 & 335.32 & 61.11 \\
    Advanced-CoLight & Full-shot  & \textbf{247.31} & 32.54 & \textbf{235.78} & 25.96 & 242.56 & 27.55 & 285.32 & 24.11 & \textbf{323.19} & 52.56 \\
    AttendLight      & Full-shot  & 280.11 & 47.59 & 250.53 & 34.72 & 251.34 & 36.42 & 288.94 & 27.84 & 338.41 & 55.58 \\
    LLMLight         & Full-shot  & 274.47 & 33.66 & 256.53 & 28.66 & 247.11 & 28.27 & 289.31 & 25.53 & 331.38 & 51.79 \\
    \midrule
    Traffic-R1       & Zero-shot  & 270.34 & \underline{27.95} & \underline{239.53} & \underline{21.11} & \underline{238.03} & \textbf{23.17} & \underline{277.83} & \textbf{15.51} & 324.11 & \underline{33.14} \\
    \textbf{Traffic-VL (ours)} & Zero-shot & \underline{269.71} & \textbf{27.82} & 239.81 & \textbf{21.05} & \textbf{237.95} & \underline{23.31} & \textbf{277.56} & \underline{15.62} & \underline{323.89} & \textbf{33.12} \\
    \bottomrule
  \end{tabular}
\end{table*}

\FloatBarrier

\end{document}